\documentclass[runningheads]{llncs}

\usepackage{eccv}

\usepackage{eccvabbrv}

\usepackage{graphicx}
\usepackage{booktabs}
\usepackage{multirow}
\usepackage{multicol}
\usepackage{marvosym}
\usepackage[table,xcdraw]{}
\usepackage[toc,page]{appendix}
\usepackage{longtable}
\usepackage{tabularx}

\usepackage[accsupp]{axessibility}  % Improves PDF readability for those with disabilities.

\usepackage[most]{tcolorbox}

\usepackage{hyperref}

\usepackage{orcidlink}

\begin{document}

\title{CLAP: Isolating Content from Style through Contrastive Learning with Augmented Prompts} 

\titlerunning{CLAP: Contrastive Learning with Augmented Prompts}

\author{Yichao Cai\orcidlink{0000-0003-1607-8948}\textsuperscript{(\Letter)}
\and Yuhang Liu\orcidlink{0000-0002-8195-9349} 
\and Zhen Zhang\orcidlink{0000-0003-2805-4396} 
\and Javen Qinfeng Shi\orcidlink{0000-0002-9126-2107}
}

\authorrunning{Y.~Cai et al.}

\institute{Australian Institute for Machine Learning, University of Adelaide, SA 5000, Australia \\
\email{\{yichao.cai,yuhang.liu01,zhen.zhang02,javen.shi\}@adelaide.edu.au}\\
% \url{http://www.springer.com/gp/computer-science/lncs}
}

\maketitle

\begin{abstract}
Contrastive vision-language models, such as CLIP, have garnered considerable attention for various downstream tasks, mainly due to the remarkable ability of the learned features for generalization. However, the features they learned often blend content and style information, which somewhat limits their generalization capabilities under distribution shifts. To address this limitation, we adopt a causal generative perspective for multimodal data and propose contrastive learning with data augmentation to disentangle content features from the original representations. To achieve this, we begin with exploring image augmentation techniques and develop a method to seamlessly integrate them into pre-trained CLIP-like models to extract pure content features. Taking a step further, recognizing the inherent semantic richness and logical structure of text data, we explore the use of text augmentation to isolate latent content from style features. This enables CLIP-like model's encoders to concentrate on latent content information, refining the learned representations by pre-trained CLIP-like models. Our extensive experiments across diverse datasets demonstrate significant improvements in zero-shot and few-shot classification tasks, alongside enhanced robustness to various perturbations. These results underscore the effectiveness of our proposed methods in refining vision-language representations and advancing the state-of-the-art in multimodal learning.
\footnote{Our code is available at \url{https://github.com/YichaoCai1/CLAP}.}

\end{abstract}

\section{Introduction}
\label{sec:intro}

Vision-language models, exemplified by CLIP \cite{radford2021learning}, have garnered substantial attention due to their exceptional generalization capabilities, achieved through the learned features, obtained by utilizing a cross-modal contrastive loss \cite{radford2021learning, jia2021scaling, li2021supervision}. However, despite being pre-trained on extensive datasets, CLIP-like models fall short in disentangling latent content information and latent style information. Consequently, they are not immune to spurious correlations, i.e., style-related information is erroneously utilized to predict task-related labels. These limitations become evident in the presence of distribution shifts or adversarial attacks where spurious correlations often change across different environments. For examples, (1) a notable dependence on specific input text prompts has been reported for zero-shot capabilities \cite{zhou2022learning, zhou2022conditional, khattak2023maple}; (2) performance decline in few-shot scenarios has been observed in few-shot learning scenarios \cite{radford2021learning, gao2024clip}; and (3) susceptibility to adversarial attacks has been explored \cite{mao2022understanding, yang2023robust, wortsman2022robust}.

\begin{figure}[tb]
    \centering
    \resizebox{\linewidth}{!}{
        \begin{subfigure}{0\linewidth}
        \phantomcaption
        \label{subfig:imgaug_lvm}
        \end{subfigure}
        
        \begin{subfigure}{0\linewidth}
        \phantomcaption
        \label{subfig:txtaug_lvm}
        \end{subfigure}

        \begin{subfigure}{0\linewidth}
        \phantomcaption
        \label{subfig:cam}
        \end{subfigure}
        
        \includegraphics{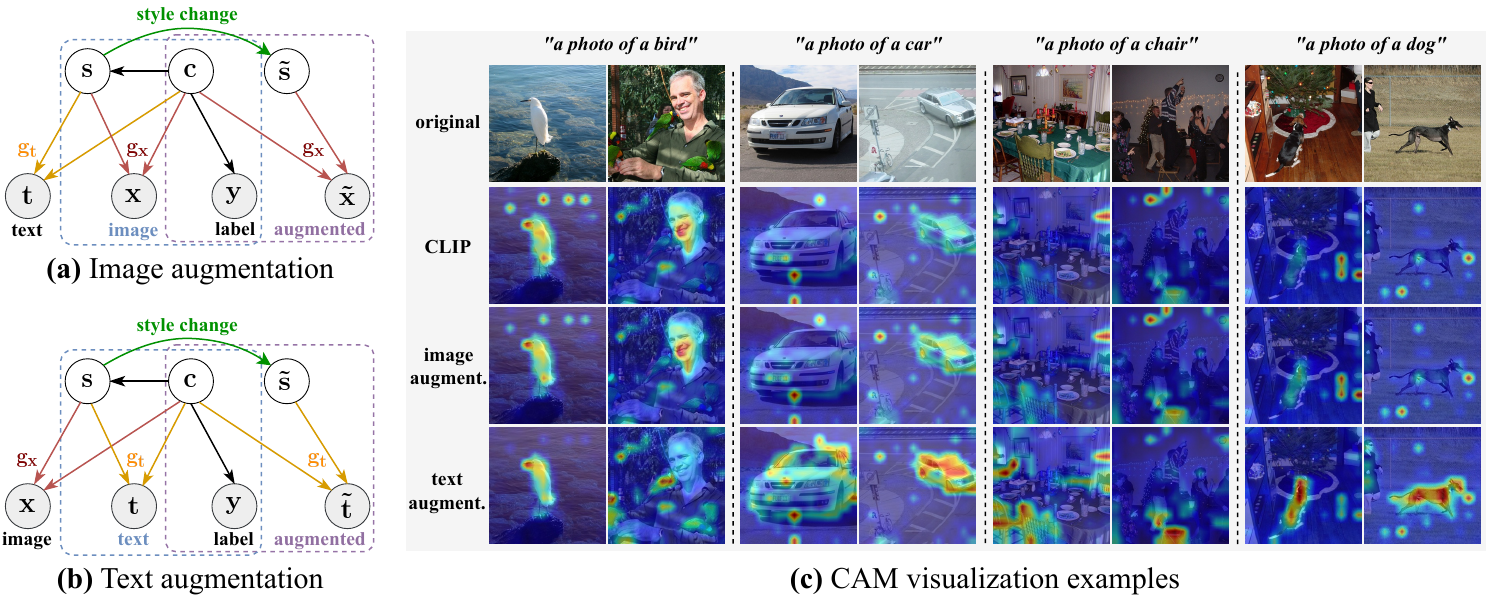}
    }
    \caption{Causal generative models of vision-language data. Image and text data are generated through distinct underlying deterministic processes, $\mathbf{g_x}$ for images and $\mathbf{g_t}$ for texts, derived from a unified latent space with latent content variables $\mathbf{c}$ and latent style variables $\mathbf{s}$. Latent content $\mathbf{c}$ exclusively determines the sample label $\mathbf{y}$. \textbf{(a)} Soft interventions on latent style variables $\mathbf{s}$ result in $\mathbf{\tilde{s}}$, subsequently generating augmented images $\mathbf{\tilde{x}}$. \textbf{(b)} Due to the same latent space, soft interventions on latent style variables $\mathbf{s}$ can also result in $\mathbf{\tilde{s}}$, producing augmented text $\mathbf{\tilde{t}}$. \textbf{(c)} A qualitative comparison of image features for zero-shot classification using "a photo of a [class]" prompts, visualized using class activation map (CAM) \cite{sMamoolerCLIPExplain}, demonstrates that while image augmentation can enhance CLIP features, the features obtained through text augmentation methods predominantly focus on the content. 
    } 
    \label{fig:causalmodel}
\end{figure}

Taking a causal perspective, this work begin with a simple yet effective method, image augmentation, to disentangle content and style information within the learned representations of CLIP-like models. This approach is inspired by recent advancements in theoretical development in causal representation learning \cite{von2021self}, which demonstrate that augmented image can be interpreted as a result of soft interventions on latent style variables, as depicted in \cref{subfig:imgaug_lvm}. Such augmentation results in a natural data pair where content information remains unchanged while style information changes. Consequently, using contrastive learning, it becomes feasible to isolate the invariant content information from the variant style information. Motivated by this theoretical advancement, we propose a practical method to incorporate image augmentation into CLIP-like models to extract content information from the original learned features. Specifically, a disentangled network is designed to fine-tune the pre-trained CLIP model by using a contrastive loss with image augmentation.

Despite the advancements made in disentangling content and style information from the original features learned by CLIP-like models through image augmentation, we recognize an inherent limitation: it is generally challenging to design adequate image augmentations to ensure all style factors change in an image. Theoretically, disentangling content and style information necessitates changes in all style factors \cite{von2021self}. However, inducing sufficient changes in latent style through image augmentation poses challenges due to the high dimensionality and complexity of style information in image data. Achieving significant style variation via artificially designed image augmentation techniques, such as transforming a photograph of a dog into a sketch while preserving content but dramatically altering style, is notably difficult.

Taking a further step, rather than relying on image augmentation, we explore the use of text augmentation to disentangle latent content and style factors. This shift is motivated by two key observations: 1) Vision and language data share the same latent space. Therefore, text augmentation can also be utilized to induce changes in latent style factors instead of image augmentation. 2) Text data inherently possesses high semanticity and logical structure, making it more amenable to property-wise manipulation compared to image data. Consequently, implementing sufficient style changes through text augmentation is more feasible than image augmentation, contributing to isolating content from style information, see \cref{subfig:cam} for visual comparison. For instance, transforming text from "a photo of a dog" to "a sketch of a dog" is straightforward in the language modality, whereas achieving a similar transformation in image data is challenging. Inspired by these observations, we posit that introducing style variations through text augmentation, as illustrated \cref{subfig:txtaug_lvm}, provides a more effective approach for learning vision-language content features than relying on image augmentation.

In summary, our contributions include: (1) Aimed at disentangling latent content and style factors to refine vision-language features of pre-trained CLIP-like models, we propose constrastive learning with data augmentation to fine tune the original features of pre-trained CLIP-like models from a causal perspective. (2) We present a novel method customized for pre-trained CLIP-like models. This method leverages a disentangled network, which is trained using contrastive learning with image augmentation, to extract latent content features from the learned features provided by image encoder of CLIP-like models. (3) We propose Contrastive Learning with Augmented Prompts (CLAP), to extract latent content features from representations of CLIP-like models. It begins by training a disentangled network using the pre-trained text encoder of CLIP-like models and text augmentation. Subsequently, the trained disentangled network is transferred to the image encoder of CLIP-like models. (4) Experiments conducted on a large real dataset demonstrate the effectiveness of the proposed image augmentation and text augmentation in terms of zero-shot and few-shot performance, as well as robustness against perturbations.

% \begin{itemize}
%     \item Aimed at disentangling latent content and style factors to refine vision-language features of pre-trained CLIP-like models, we propose constrastive learning with data augmentation to fine tune the original features of pre-trained CLIP-like models from a causal perspective.

%     \item We present a novel method customized for pre-trained CLIP-like models. This method leverages a disentangled network, which is trained using contrastive learning with image augmentation, to extract latent content features from the learned features provided by image encoder of CLIP-like models.
    
%     \item We propose Contrastive Learning with Augmented Prompts (CLAP), to extract latent content features from representations of CLIP-like models. It begins by training a disentangled network using the pre-trained text encoder of CLIP-like models and text augmentation. Subsequently, the trained disentangled network is transferred to the image encoder of CLIP-like models.

%      \item Experiments conducted on a large real dataset demonstrate the effectiveness of the proposed image augmentation and text augmentation in terms of zero-shot and few-shot performance, as well as robustness against perturbations.
% \end{itemize}

\section{Related Work}
\label{sec:relate}

\subsubsection{Contrastive Vision-Language Models}
Using a cross-modal contrastive loss, CLIP \cite{radford2021learning} revolutionarily introduced a scalable contrastive vision-language model by leveraging a large corpus of internet-sourced image-text pairs, demonstrating unprecedented zero-shot learning capabilities and exceptional generalization ability across datasets and supporting numerous downstream tasks \cite{ramesh2021zero}. ALIGN \cite{jia2021scaling} expanded the scale of contrastive vision-language modeling, training on up to one billion image-text pairs while integrating the vision transformer's self-attention mechanism \cite{dosovitskiy2020image}, which further enhanced performance. Despite their successes, CLIP-like models exhibit sensitivity to input text prompts \cite{zhou2022learning, khattak2023maple}, leading to variable performance across different prompts. Efforts to mitigate this prompt sensitivity through prompt learning and engineering \cite{zhou2022learning, khattak2023maple, cho2023promptstyler, zhou2022conditional, ge2022domain} focus on customizing prompts for specific tasks but do not fundamentally enhance CLIP's representations. Furthermore, CLIP-like models are vulnerable to adversarial attacks \cite{carlini2021poisoning, fort2022adversarial}, with current strategies \cite{mao2022understanding, yang2023robust} involving adversarial-natural image pairs to improve resilience. Our work diverges from task-specific approaches by aiming to enhance CLIP's representations from a disentanglement perspective, addressing the aforementioned issues inherent in CLIP-like models.

\subsubsection{Disentangled Representation Learning}
Aimed at segregating intrinsic latent factors in data into distinct, controllable representations, disentangled representation learning benefits various applications \cite{yang2021causalvae, sanchez2020learning, li2021disentangled}. Specifically, in classification tasks, it's shown that enhancing the model's performance and robustness against data distribution perturbations can be achieved by more effectively disentangling invariant content variables, without needing to identify all intrinsic latent variables completely \cite{kong2022partial, liu2023identifiable,liu2022identify,liu2024identifiable}. Within single modalities, studies such as \cite{zimmermann2021contrastive} have illustrated that contrastive learning \cite{chen2020simple, grill2020bootstrap, he2020momentum} can potentially reverse the data generative process, aiding in the separation of representations. Furthermore, \cite{von2021self} suggest that image augmentation can help isolate content variables from the latent space through significant stylistic changes. \cite{hong2022itmix} employs mixture techniques for data augmentation, enabling more abundant cross-modal matches. Diverging from these methods, our approach focuses on employing text augmentation to disentangle latent content variables, introducing a unique approach to learn refined vision-language representations.

\section{A Causal Generative Model for Multi-Modal Data}
\label{sec:causality}

To understand pretrained CLIP-like models, we investigate the underlying causal generative process for vision-language data. We consider the following causal generative model as depicted in \cref{fig:causalmodel}. In the proposed model, the shared latent space ruling vision and language data is divided into two distinct sub-spaces: one corresponding to the latent content variables $\mathbf{c}$ and the other to the latent style variables $\mathbf{s}$. The latent content variables are posited to determine the object label $\mathbf{y}$, a relationship corroborated by prior studies \cite{kong2022partial, liu2022identifying, mahajan2021domain}. Furthermore, to elucidate the correlation between the latent style variable $\mathbf{s}$ and the object variable $\mathbf{y}$, our model incorporates the premise that the latent content variable $\mathbf{c}$ causally influences the latent style variable $\mathbf{s}$, in concordance with the principles of causal representation learning highlighted in recent literature \cite{daunhawer2023identifiability, liu2022identifying, von2021self}. Additionally, considering the diversity between image data and text data, where information in image data is typically much more details while information in text data tends to be more logically structured nature, we posit distinct causal mechanisms for the generation processes. Our causal generative model is formulated as following structural causal models \cite{bollen1989structural}:
\begin{equation}
    \mathbf{s} := \mathbf{g_s(c)},\quad \mathbf{x} := \mathbf{g_x(c, s)},\quad \mathbf{t} := \mathbf{g_t(c, s)},\quad \mathbf{y} := \mathbf{g_y(c)}. 
    \label{eq:lvm}
\end{equation}
In \cref{eq:lvm}, the style variable $\mathbf{s}$ is causally influenced by the content via $\mathbf{g_s}$; $\mathbf{x}$ and $\mathbf{t}$ denote visual and textual data, respectively. Both visual and textual data are causally produced by the shared latent variables $\mathbf{c}$ and $\mathbf{s}$ through distinct, reversible generative processes: $\mathbf{g_x}$ for images and $\mathbf{g_t}$ for text data, respectively. The label $\mathbf{y}$ of a sample is exclusively determined by the content variable $\mathbf{c}$ via $\mathbf{g_y}$.For simplicity, exogenous noises are implicitly assumed but not explicitly represented in the causal generative model's formulation, aligning with the common understanding that each latent variable is influenced by exogenous noise. 
 
Recent seminal work in \cite{von2021self} has demonstrated that the latent content variable $\mathbf{c}$ can be identified up to block identifiability (i.e., $\mathbf{c}$ can be isolated from style variable $\mathbf{s}$), by requiring all latent style variables to change (e.g., soft interventions on all latent style variables). This change can be achieved through image augmentation, i.e., the augmented image $\mathbf{\tilde{x}}$ can be interpreted as a generative result of $\mathbf{\tilde{s}}$, which is produced through soft interventions on original latent style variables $\mathbf{s}$. Despite such theoretical advancement, the practical implementation of this theoretical result within CLIP-like models remains unclear. In this study, we propose a practical method to disentangle content and style information within CLIP-like models by employing image augmentation, as detailed in \cref{sec: imgaug}. Moreover, we recognize that implementing sufficient changes on all latent style variables $\mathbf{s}$ through text augmentation is more feasible than image augmentation, due to high semanticity and logical structure in text data, we delve into the use of text augmentation to separate content information from style information, as discussed in \cref{sec: textaug}.

\section{Isolating Content from Style with Data Augmentation}
In this section, we propose the employment of data augmentation to extract content information from the learned features in pre-trained CLIP-like models. Essentially, data augmentation facilitates the alteration of style factors while preserving content factors. Consequently, leveraging contrastive learning enables the segregation of content information from style information. We delve into two distinct forms of data augmentation, namely image augmentation (\cref{sec: imgaug}) and text augmentation (\cref{sec: textaug}).

\subsection{Isolating Content from Style with Augmented Images} \label{sec: imgaug}
While recent studies (von et al., 2021) have offered assurance regarding the disentanglement of content and style through contrastive learning with data augmentation, it remains unclear how these theoretical findings can be applied to the realm of vision-language models. We convert the theoretical findings into CLIP-like models in the following. The theoretical findings suggest using InfoNCE loss \cite{oord2018representation} to extract content information, as outlined below:
\begin{equation}
    \mathcal{L}{(\mathbf{f}; \{\mathbf{x}_i, \mathbf{\tilde{x}}_i\}_{i=1}^{b}, \tau)} = 
        -\frac{1}{b}\sum\nolimits_{i=1}^{b} \mathrm{log}
        \frac{\mathrm{exp}\left[\langle\mathbf{f}(\mathbf{x}_i), \mathbf{f}(\mathbf{\tilde{x}}_i)\rangle / \tau\right]}{\sum_{j=1}^{b} \mathrm{exp}\left[\langle\mathbf{f}(\mathbf{x}_i), \mathbf{f}(\mathbf{\tilde{x}}_j)\rangle / \tau\right]},
    \label{eq:infonce}
\end{equation}
where $\{\mathbf{x}_i\}_{i=1}^{b}$ represents a batch of $b$ samples from the training dataset, $\mathbf{f}(\mathbf{x}_i)$ denotes sample $\mathbf{x}_i$'s features through model $\mathbf{f}$, $\mathbf{\tilde{x}}_i$ is the augmented counterpart of $\mathbf{x}_i$, and $\mathrm{\langle\mathbf{z}_1, \mathbf{z}_2\rangle}$ represents the cosine similarity between two feature vectors, $\mathbf{z}_1$ and $\mathbf{z}_2$, and $\tau$ represents the temperature parameter influencing the loss.

\begin{figure}[tb]
    \centering
  \resizebox{\linewidth}{!}{
        \begin{subfigure}{0.3\linewidth}
            \includegraphics[width=1\linewidth]{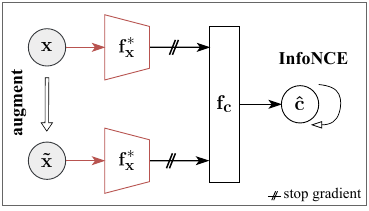}
        \caption{Image augmentation}
        \label{subfig:trainimgaug}
        \end{subfigure}
        
        \begin{subfigure}{0.3\linewidth}
            \includegraphics[width=1\linewidth]{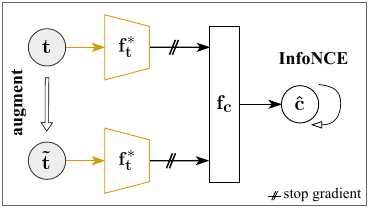}
        \caption{Text augmentation}
        \label{subfig:train}
        \end{subfigure}
        
        \begin{subfigure}{0.3\linewidth}
            \includegraphics[width=1\linewidth]{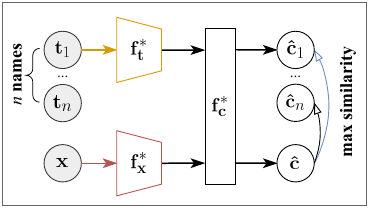}
        \caption{Zero-shot inference}
        \label{subfig:inference}
        \end{subfigure}
    }
  \caption{Refining CLIP through data augmentation. (a) Training involves a disentangled network $\mathbf{f_c}$, utilizing contrastive loss on original and augmented image pairs $\mathbf{x}$ and $\mathbf{\tilde{x}}$, with CLIP's image encoder $\mathbf{f^*_x}$ holding frozen gradients. (b) More efficient content feature learning is achieved through contrastive learning with augmented text prompts $\mathbf{t}$ and $\mathbf{\tilde{t}}$, using the fixed text encoder $\mathbf{f^*_t}$ of CLIP. (c) Inference stage: The trained disentangled network $\mathbf{f^*_c}$ integrates with CLIP's text and image encoders, $\mathbf{f^*_t}$ and $\mathbf{f^*_x}$, to enable zero-shot inference for an input image $\mathbf{x}$ and class names $\mathbf{t}_1$ to $\mathbf{t}_n$.}
  \label{fig:framework}
\end{figure}

We extend it to refine pre-trained vision-language models, utilizing contrastive learning with augmented images (hereinafter referred to as "\textbf{Im.Aug}"). As illustrated in \cref{subfig:trainimgaug}, we train a disentangled network on top of CLIP's pre-trained image encoder. To enhance training efficiency and the usability of the proposed method, we freeze the pre-trained image encoder. Based on an InfoNCE loss, the learning objective of Im.Aug is formulated as follows:
\begin{equation}
    \mathbf{f^*_c} = \mathop{\mathrm{argmin}}_{\mathbf{f_c}} \mathop{\mathbb{E}}_{\{\mathbf{x}_i\}_{i=1}^b \in \mathcal{D}_\mathbf{x}}
    \mathcal{L}(\mathbf{f_c \circ f^*_x};  \{\mathbf{x}_i, \mathbf{\tilde{x}}_i\}_{i=1}^{b}, \tau),
    \label{eq:loss_imaug}
\end{equation}
where $\mathcal{D}_\mathbf{x}$ denotes the training image dataset and $b$ represents the batch size, $\mathbf{f_c}$ is the disentangled network undergoing training. The pre-trained CLIP image encoder is represented by $\mathbf{f}^*_\mathbf{x}$, with the asterisk "*" signifying that the model weights remain fixed. The variable $\mathbf{x}_i$ refers to an image sampled from $\mathcal{D}_\mathbf{x}$, and $\mathbf{\tilde{x}}_i$ is its augmented view. 

The composition of the training dataset $\mathcal{D}_\mathbf{x}$, the image augmentation techniques used, the structure of the disentangled network $\mathbf{f_c}$, and the utilization of $\mathbf{f^*_c}$ post-training are detailed in the following subsections.

\subsubsection{Data Synthesis and Image Augmentation}
To generate training image data, we combine class names with various image and object attributes to create text prompts for each class. Using a stable diffusion model \cite{rombach2022high}, we produce synthetic images that comprise our training dataset $\mathcal{D}\mathbf{x}$. The creation of template prompts for stable diffusion is based on attributes such as object size, color, image type, and art style. As detailed in \cref{tab:genprompts}, the attributes include 10 colors and 3 sizes for objects, and 8 types and 2 art styles for images. By assembling these attributes into prompts like "a [art style] [image type] of a [object size] [object color] [class]", we generate 480 unique texts for each class, from which one image per prompt is synthesized. Further details on image synthesis and examples are available in \cref{subsec:sd-gen}.
\begin{table}[tb]
  \caption{Template-based prompts. Attributes used to generate text prompts follow the structured format "a [art style] [image type] of a [object size] [object color] [class]", where "[class]" represents the class names.}
  \label{tab:genprompts}
  \centering
    \resizebox{\linewidth}{!}{
        \begin{tabular}{@{}cccc@{}}
        \toprule
        \textbf{Object Color}                             & \textbf{Object Size}              & \textbf{Image Type}                                   & \textbf{Art Style} \\ \midrule
        \multicolumn{1}{c|}{yellow, green, black,} & \multicolumn{1}{c|}{large,}       & \multicolumn{1}{c|}{~painting, cartoon, infograph,~}    & realistic,          \\
        \multicolumn{1}{c|}{blue, multicolored, orange,~}      & \multicolumn{1}{c|}{small,}       & \multicolumn{1}{c|}{sketch, photograph, clipart,} & ~impressionistic    \\
        \multicolumn{1}{c|}{red, white, brown, purple}  & \multicolumn{1}{c|}{~normal sized~~} & \multicolumn{1}{c|}{mosaic art, sculpture}            &                    \\ \bottomrule
        \end{tabular}
    }
\end{table}
For the image augmentation procedures, we adopt techniques commonly used in contrastive learning practice \cite{chen2020simple, von2021self, chen2021exploring}, specifically random cropping and color distortion.

\subsubsection{Disentangled Network Structure} 
\begin{figure}[tb]
    \centering
    \includegraphics[width=0.56\linewidth]{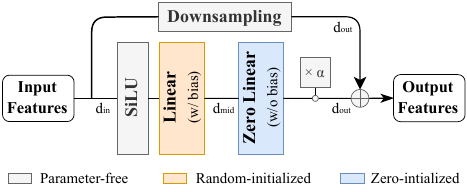}
    \caption{Structure of the disentangled network. The architecture encompass a residual block featuring a zero-initialized, bias-free linear layer to commence optimization from the input feature space.  When the input and output dimension differ, a downsampling operation is utilized to achieve alignment. During inference, a scalar parameter $\alpha$ balance the main branch and input features before combination.}
    \label{fig:mlp}
\end{figure}

Since the training process is based on CLIP's pre-trained lower-dimensional features, our disentangled network adopts a multi-layer perceptron (MLP) architecture. To fully benefit from the pre-trained CLIP text encoder, we construct a residual MLP featuring a zero-initialized projection, acting as the disentangled network, as depicted in \cref{fig:mlp}. This design enables learning directly from the pre-trained representation space, avoiding a random starting point, inspired by ControlNet's zero-conv operation \cite{zhang2023adding}, which we adapt to a zero-linear operation within our residual MLP.

Within this architecture, the main branch includes a zero-initialized, bias-free linear layer positioned subsequent to the combination of a SiLU activation and a normally initialized linear layer. Conventionally, the dimensions of features before the initial linear layer, situated between the first and second linear layers, and following the second linear layer, are named as the input $d_{in}$, latent $d_{mid}$, and output $d_{out}$ dimensions, respectively. To rectify any mismatches between the input and output dimensions, the network employs nearest-neighbor downsampling within the shortcut path, thereby ensuring both alignment and the preservation of sharpness for the input features. During the inference stage, a weighting parameter $\alpha > 0$ is introduced to modulate the portion of features emanating from the main branch before their integration with the input features, whereas this parameter remains constant at 1 throughout the training phase.

\subsubsection{Inference} After training, the disentangled network $\mathbf{f^*_c}$ is utilized following CLIP's image encoder to extract visual content features. Moreover, given that vision-language data generation is rooted in a unified latent space, as depicted in \cref{sec:causality}, $\mathbf{f^*_c}$ can be seamlessly integrated with CLIP's image and text encoders to enhance zero-shot capabilities. As shown in \cref{subfig:inference}, for an image $\mathbf{x}$, the operation is formulated as the composition function $\mathbf{f^*_c \circ f^*_x(x)}$, and similarly, for a text $\mathbf{t}$, as $\mathbf{f^*_c \circ f^*_t(t)}$. This integration preserves CLIP's zero-shot functionality while achieving refined features through the improved disentanglement of content.

\subsection{Isolating Content from Style with Augmented Prompts}
\label{sec: textaug}
Despite progress in disentangling content and style via image augmentation, adequately altering all style factors in an image remains challenging due to the high dimensionality and complexity of style information in images. Achieving substantial style changes through augmentation, essential for complete disentanglement \cite{von2021self}, is difficult with existing image augmentation techniques. On the contrary, text data inherently possesses high semanticity and logical structure, making it more amenable to property-wise manipulation compared to image data. To further exploring the disentanglement of content, we propose Contrastive Learning with Augmented Prompts (\textbf{CLAP}). 

As depicted in \cref{subfig:train}, CLAP employs an InfoNCE loss to train a disentangled network atop CLIP's pre-trained text encoder, keeping the encoder's gradients fixed, similar to Im.Aug. Leveraging the simpler structure of text, the template-based prompts previously utilized for synthesizing images now serve as the training text dataset, denoted by $\mathcal{D}_\mathbf{t}$. Utilizing the same disentangled network as in Im.Aug, the learning objective of CLAP is outlined as follows:
\begin{equation}
    \mathbf{f^*_c} = \mathop{\mathrm{argmin}}_{\mathbf{f_c}} \mathop{\mathbb{E}}_{\{\mathbf{t}_i\}_{i=1}^b \in \mathcal{D}_\mathbf{t}}
    \mathcal{L}(\mathbf{f_c \circ f^*_t};  \{\mathbf{t}_i, \mathbf{\tilde{t}}_i\}_{i=1}^{b}, \tau) + 
    \lambda \mathcal{L}(\mathbf{f_c \circ f^*_t};  \{\mathbf{t}^{c}_i, \mathbf{\tilde{t}}_i\}_{i=1}^{b},1),
    \label{eq:claploss}
\end{equation} 
 where $\mathbf{f}^*_\mathbf{t}$ denotes the pre-trained CLIP text encoder. The term $\mathbf{t}_i$ references a text prompt from $\mathcal{D}\mathbf{t}$, and $\mathbf{\tilde{t}}_i$ represents its augmented view, produced via prompt augmentation techniques. On the equation's right side, $\mathbf{t}_i^c$ specifies the class name associated with the text prompt $\mathbf{t}_i$. This strategy aims to enhance variations between prompt pairs, especially in cases where the text dataset $\mathcal{D}_\mathbf{t}$ has a very limited number of samples. Here, $\lambda$ serves for adjusting the second term's importance in the total loss function. All other symbols in \cref{eq:claploss} maintain their definitions as described earlier.

After training, the learned disentangled network is seamlessly integrated with both of CLIP's encoders to extract content representations, as depicted in \cref{subfig:inference}.

\subsubsection{Prompt Augmentation} To ensure text prompts undergo stylistic changes without compromising their content, we have developed specific augmentation techniques for synthetic text prompts. Drawing inspiration from Easy Data Augmentation (EDA) techniques \cite{wei2019eda}, we adapted the Random Deletion (RD) and Random Swap (RS) techniques from EDA, customizing them to suit our prompt structure. To avoid inadvertently altering the content by introducing new object names or changing the core idea of a text prompt, our augmentation methods do not include random word insertions or replacements. Our primary augmentation techniques are Object Size Deletion (OSD), Object Color Deletion (OCD), Image Type Deletion (ITD), Art Style Deletion (ASD), and Swapping Prompt Order (SPO), each applied with a certain probability, as detailed in \cref{tab:augtech}. Additionally, for down-stream datasets with few categories, to rich the population of training samples, we use an additional augmentation, named IGN (Inserting Gaussian Noise). Following the initializing protocol of prompt learning methods \cite{zhou2022learning, zhou2022conditional}, we insert a zero-mean Gaussian noise with 0.02 standard deviation  with a noise length equals to 4, to the tokenized prompts. 

% In practice, these augmentation techniques can be combined in various ways to achieve optimal functionality. 
Intuitively, these prompt augmentation methods parallel random masking techniques used in image augmentation \cite{he2022masked, chen2020gridmask}. However, prompt augmentations are more effective and precise than their image counterparts. This effectiveness arises because prompt augmentations can specifically target and eliminate a particular style element without impacting the content, whereas image masking, operating at the pixel or patch level, might inadvertently damage content information or lead to insufficient style changes.

\begin{table}[tb]
  \caption{Prompt augmentation techniques. Various augmented views are generated from an original text prompt using specific augmentation techniques: OSD (Object Size Deletion), OCD (Object Color Deletion), ITD (Image Type Deletion), ASD (Art Style Deletion), and SPO (Swapping Prompt Order).}
  \label{tab:augtech}
  \centering
    \resizebox{\linewidth}{!}{
        \begin{tabular}{@{}cccccc@{}}
        \toprule
        \textbf{Original}                  & \textbf{OSD}                       & \textbf{OCD}                        & \textbf{ITD}                       & \textbf{ASD}                       & \textbf{SPO}       \\ \midrule
        \multicolumn{1}{c|}{a realistic}   & \multicolumn{1}{c|}{a realistic}   & \multicolumn{1}{c|}{a realistic}    & \multicolumn{1}{c|}{a realistic}   & \multicolumn{1}{c|}{a painting}    & a large red     \\
        \multicolumn{1}{c|}{painting of a~~} & \multicolumn{1}{c|}{~painting of a~~} & \multicolumn{1}{c|}{~painting of a~~} & \multicolumn{1}{c|}{of a}          & \multicolumn{1}{c|}{of a}          & car in a               \\
        \multicolumn{1}{c|}{large red car} & \multicolumn{1}{c|}{red car}       & \multicolumn{1}{c|}{large car}      & \multicolumn{1}{c|}{~large red car~~} & \multicolumn{1}{c|}{~large red car~~} & ~realistic painting \\ \bottomrule
        \end{tabular}
    }
\end{table}

\section{Experiments}
\label{sec:expriment}
We conduct three primary experiments to assess our method: (1) zero-shot evaluation with diverse prompts to gauge zero-shot performance and its robustness to prompt perturbations; (2) linear probe tests on few-shot samples to evaluate the efficacy of the learned representations in few-shot settings; and (3) adversarial attack assessments on zero-shot and one-shot classifiers to determine their resistance to adversarial threats. We further conduct an ablative study on hyper-parameters, explore the impact of different prompt augmentation combinations and various sources of training prompts on CLAP's performance, and replicate experiments across different CLIP model sizes.

\subsection{Experimental Setup}
\textit{Implementation.} Im.Aug and CLAP are implemented using the ViT-B/16 CLIP model and executed on an NVIDIA RTX 3090 GPU. To ensure reproducibility, the random seed for all stochastic processes is fixed at 2024. More information on implementation details is provided in \cref{subsec:implementdetail}.

\textit{Datasets.} CLAP is assessed across four multi-domain datasets to examine its performance in varied environments: PACS \cite{li2017deeper} (4 domains, 7 categories), VLCS \cite{albuquerque2019generalizing} (4 domains, 5 categories), OfficeHome \cite{rahman2019multi} (4 domains, 65 categories), and DomainNet \cite{peng2019moment} (6 domains, 345 categories). For conciseness, we present average results across the domains for each dataset. Detailed experimental outcomes for each domain within these datasets are provided in \cref{subsec:vitb-details}.

\textit{Compute efficiency.} CLAP demonstrates faster convergence and shorter training times compared to Im.Aug. For CLAP, training on the PACS and VLCS datasets is completed in roughly 11 minutes, OfficeHome in approximately 14 minutes, and DomainNet in about 47 minutes. In contrast, Im.Aug requires around 16 minutes for PACS and VLCS, 50 minutes for OfficeHome, and 3.3 hours for DomainNet. Both Im.Aug and CLAP maintain CLIP's inference efficiency due to the disentangled network's efficient two-layer MLP structure.

\subsection{Main Results}
\subsubsection{Zero-Shot Performance}
To assess zero-shot capabilities, CLAP undergoes evaluation using three specific fixed prompts: ZS(C), utilizing only the class name within "[class]"; ZS(PC), with the format "a photo of a [class]"; and ZS(CP), structured as "a [class] in a photo". To thoroughly examine zero-shot proficiency, a dynamic prompt, ZS(NC), formatted as "[noise][class]", is also used, where "[noise]" signifies the introduction of Gaussian noise characterized by a mean of 0 and a standard deviation of 0.02.

As presented in \cref{tab:zeroshot}, CLAP surpasses both CLIP and Im.Aug across all evaluated prompts for every dataset. Unlike the uniform enhancement in zero-shot performance CLAP achieves over CLIP, Im.Aug displays inconsistent results. A closer examination reveals CLAP's superiority over CLIP is especially significant for the dynamic ZS(NC) prompt. This demonstrates CLAP's effectiveness in significantly improving zero-shot performance compared to the original CLIP representations.

\begin{table}[tb]
  \caption{Zero-shot results across three distinct prompts: "C" for "[class]", "CP" for "a [class] in a photo", "PC" for "a photo of a [class]", and a dynamic prompt "NC" for "[noise][class]" showcase that CLAP consistently outperforms CLIP's zero-shot performance across all datasets, whereas image augmentation exhibits mixed outcomes.
  }
  \label{tab:zeroshot}
  \centering
    % \resizebox{\linewidth}{!}{
    \begin{tabular}{@{}ccccccc@{}}
    \toprule
    \multirow{2}{*}{Prompt} & \multirow{2}{*}{Method} & \multicolumn{5}{c}{Zero-shot performance,  avg. top-1 acc. (\%) (↑)}          \\ \cmidrule(l){3-7} 
                                       &                         & ~PACS~~          & ~VLCS~~          & Off.Home    & ~Dom.Net     & Overall       \\ \midrule
    \multirow{3}{*}{ZS(C)}             & CLIP                    & 95.7          & 76.4          & 79.8          & 57.8          & 77.4          \\
                                       & Im.Aug                  & 96.5          & 79.5          & 77.0          & 51.5          & 76.1          \\
                                       & CLAP                    & \textbf{97.2} & \textbf{82.6} & \textbf{81.0} & \textbf{58.7} & \textbf{79.9} \\ \midrule
    \multirow{3}{*}{ZS(CP)}            & CLIP                    & 95.2          & 82.0          & 79.5          & 57.0          & 78.4          \\
                                       & Im.Aug                  & 96.3          & 82.9          & 75.8          & 50.7          & 76.4          \\
                                       & CLAP                    & \textbf{97.3} & \textbf{83.4} & \textbf{80.5} & \textbf{58.0} & \textbf{79.8} \\ \midrule
    \multirow{3}{*}{ZS(PC)}            & CLIP                    & 96.1          & 82.4          & 82.5          & 57.7          & 79.7          \\
                                       & Im.Aug                  & 96.5          & 83.0          & 78.6          & 51.6          & 77.4          \\
                                       & CLAP                    & \textbf{97.2} & \textbf{83.4} & \textbf{83.0} & \textbf{59.0} & \textbf{80.6} \\ \midrule
    \multirow{3}{*}{ZS(NC)}            & CLIP                    & 90.8          & 68.3          & 71.5          & 51.0          & 70.4          \\
                                       & Im.Aug                  & 94.8          & 73.1          & 67.5          & 44.0          & 69.9          \\
                                       & CLAP                    & \textbf{97.2} & \textbf{81.0} & \textbf{73.5} & \textbf{52.6} & \textbf{76.1} \\ 
    \bottomrule
    \end{tabular}
    % }

\end{table}

In assessing the model's robustness to prompt perturbations, we examine the variances in zero-shot performance across different prompts by analyzing the range ($R$) and standard deviation ($\delta$) of results derived from ZS(C), ZS(CP), and ZS(PC). Additionally, we investigate the decline ($\Delta_{(NC)}$) in performance from ZS(C) to ZS(NC) as a broad indicator of resilience to noisy prompts. 

As presented in \cref{tab:variance}, CLAP significantly reduces the variance in zero-shot performance across various testing prompts, evidenced by markedly lower values of $\delta$ and $R$, and a less pronounced decrease in performance with a noised prompt, in contrast to Im.Aug and the baseline representations of CLIP. Although Im.Aug aids in reducing performance variance to some extent, its efficacy is notably inferior to that of CLAP. These findings highlight CLAP's enhanced robustness in maintaining consistent zero-shot performance across a diverse array of prompts.

\begin{table}[tb]
  \caption{
CLAP more effectively reduces zero-shot performance variance across prompts than image augmentation, with $R$ and $\delta$ indicating the range and standard deviation for ZS(C), ZS(CP), and ZS(PC). The decrease $\Delta_{(NC)}$ from ZS(C) to ZS(NC) highlights CLAP's enhanced robustness against prompt perturbations.
  }
  \label{tab:variance}
  \centering
    % \resizebox{\linewidth}{!}{
    \begin{tabular}{@{}ccccccc@{}}
    \toprule
    \multirow{2}{*}{Metric}  & \multirow{2}{*}{Method} & \multicolumn{5}{c}{Performance variance,   avg. top-1 acc. (\%) (↓)}  \\ \cmidrule(l){3-7} 
                             &                         & ~PACS~~         & ~VLCS~~         & Off.Home   & ~Dom.Net    & Overall      \\ \midrule
    \multirow{3}{*}{$R$}       & CLIP                    & 0.9          & 6.1          & 3.1          & \textbf{0.8} & 2.7          \\
                             & Im.Aug                  & \textbf{0.1}          & 3.6          & 2.8          & 0.9          & 1.9          \\
                             & CLAP                    & \textbf{0.1} & \textbf{0.8} & \textbf{2.5} & 1.0 & \textbf{1.1} \\ \midrule
    \multirow{3}{*}{$\delta$}       & CLIP                    & 0.4          & 2.8          & 1.4          & \textbf{0.4} & 1.2          \\
                             & Im.Aug                  & 0.1          & 1.7          & 1.2          & \textbf{0.4}          & 0.8          \\
                             & CLAP                    & \textbf{0.0} & \textbf{0.4} & \textbf{1.1} & \textbf{0.4} & \textbf{0.5} \\ \midrule
    \multirow{3}{*}{$\Delta_{(NC)}$} & CLIP                    & 4.9          & 8.1          & 8.3          & 6.8          & 7.0          \\
                             & Im.Aug                  & 1.6          & 6.4          & 9.5          & 7.5          & 6.3          \\
                             & CLAP                    & \textbf{0.0} & \textbf{1.6} & \textbf{7.5} & \textbf{6.1} & \textbf{3.8} \\ \bottomrule
    \end{tabular}
    % }
\end{table}

\subsubsection{Few-Shot Performance} 
\begin{figure}[tb]
  \centering
    \resizebox{1\linewidth}{!}{
    \includegraphics[width=\linewidth]{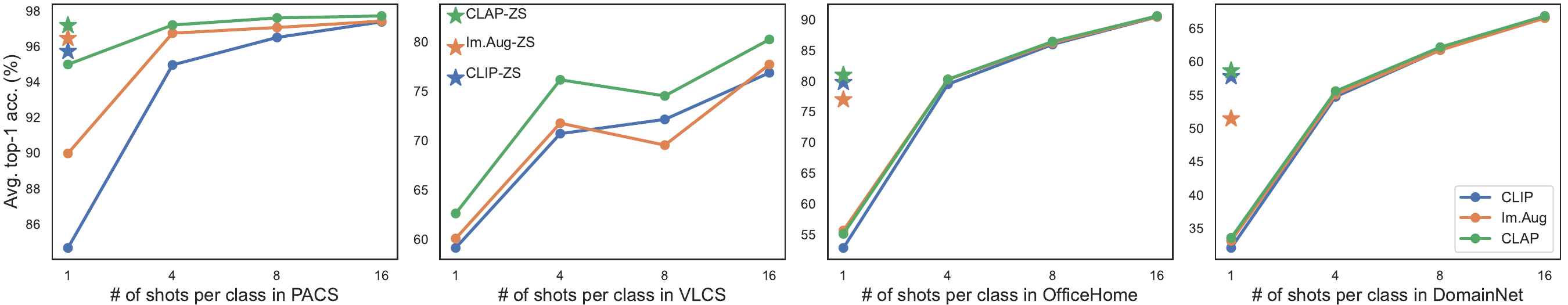}
    }
  \caption{Few-shot linear probe comparisons of image-encoder features show that CLAP enhances CLIP's few-shot performance more effectively than Im.Aug. In the accompanying figure, "ZS" indicates the zero-shot performance using a "[class]" prompt.}
  \label{fig:fewshot}
\end{figure}

We conduct evaluations of 1-shot, 4-shot, 8-shot, and 16-shot linear probes across each domain within the four datasets. As illustrated in \cref{fig:fewshot}, CLAP significantly outperforms both CLIP and Im.Aug in few-shot learning scenarios. Notably, in the 1-shot setting CLAP achieves performance improvements over the linear-probe CLIP model by margins of +10\%, +3.5\%, +2.5\%, and +1.5\% on the PACS, VLCS, OfficeHome, and DomainNet datasets, respectively. These improvements are especially significant in comparison to the gains observed with Im.Aug counterparts, underpinning CLAP's efficacy in few-shot scenarios.  For detailed quantitative results, please refer to \cref{subsec:few-shot}.

\subsubsection{Adversarial Performance}
To assess adversarial robustness, zero-shot (ZS(C)) and one-shot classifiers are evaluated against prominent adversarial attack methods, such as FGSM \cite{goodfellow2015explaining}, PGD \cite{madry2018towards}, and CW \cite{carlini2017towards}, by generating adversarial samples for testing. For FGSM, 1 adversarial iteration is employed, whereas for PGD and CW, 20 iterations are used, all with an epsilon of 0.031. As indicated in \cref{tab:adversarial}, classifiers using CLAP representations exhibit greater resilience to these adversarial attacks than those using CLIP representations. Across the four datasets, CLAP's zero-shot and 1-shot classifiers surpass CLIP by margins of +2.3 and +8.5 percentage points against FGSM, +1.0 and +11.7 percentage points against PGD-20, and +1.1 and +2.3 percentage points against CW-20, respectively. These gains are generally larger than the corresponding changes of -0.8 and +4.5 percentage points against FGSM, +0.3 and +6.2 percentage points against PGD-20, and 0.0 and +1.3 percentage points against CW-20 achieved by Im.Aug. The result suggests that CLAP efficiently enhances robustness against adversarial attacks in both zero-shot and one-shot scenarios.

\begin{table}[tb]
    \caption{Average top-1 accuracy (\%) under FGSM, PGD-20, and CW-20 adversarial attacks for zero-shot ZS(C) and 1-shot classifiers. Both refinement methods improve robustness in several settings, with CLAP attaining the highest overall accuracy.}
    \label{tab:adversarial}
    \centering
    \resizebox{\linewidth}{!}{
    \begin{tabular}{@{}ccccccccccccccc@{}}
    \toprule
    \multirow{3}{*}{Setting} & \multirow{3}{*}{Method} & \multicolumn{12}{c}{Top-1 accuracy under adversarial attack (\%) (\(\uparrow\))} & \\
    \cmidrule(l){3-15}
     & & \multicolumn{4}{c|}{FGSM} & \multicolumn{4}{c|}{PGD-20} & \multicolumn{4}{c|}{CW-20} & \multirow{2}{*}{Avg.} \\
     & & PACS & VLCS & O.H. & \multicolumn{1}{c|}{D.N.} & PACS & VLCS & O.H. & \multicolumn{1}{c|}{D.N.} & PACS & VLCS & O.H. & \multicolumn{1}{c|}{D.N.} & \\ \midrule
    \multirow{3}{*}{ZS(C)}
      & CLIP   & 86.8 & 65.6 & 57.9 & \multicolumn{1}{c|}{43.6} & 29.1 & 2.0 & 10.1 & \multicolumn{1}{c|}{10.7} & 27.4 & 1.5 & 7.4 & \multicolumn{1}{c|}{7.6} & 29.1 \\
      & Im.Aug & 88.0 & 69.6 & 55.1 & \multicolumn{1}{c|}{37.9} & \textbf{31.3} & 2.1 & 10.4 & \multicolumn{1}{c|}{9.0} & 29.4 & 1.7 & 7.0 & \multicolumn{1}{c|}{5.8} & 28.9 \\
      & CLAP   & \textbf{88.7} & \textbf{71.9} & \textbf{58.5} & \multicolumn{1}{c|}{\textbf{44.2}} & 30.8 & \textbf{3.2} & \textbf{10.6} & \multicolumn{1}{c|}{\textbf{11.2}} & \textbf{29.8} & \textbf{2.3} & \textbf{8.1} & \multicolumn{1}{c|}{\textbf{8.0}} & \textbf{30.6} \\ \midrule
    \multirow{3}{*}{1-shot}
      & CLIP   & 66.7 & 45.2 & 34.3 & \multicolumn{1}{c|}{22.5} & 34.8 & 16.0 & 5.6 & \multicolumn{1}{c|}{11.3} & 18.9 & 0.7 & 4.5 & \multicolumn{1}{c|}{3.2} & 22.0 \\
      & Im.Aug & 79.4 & 46.6 & \textbf{37.1} & \multicolumn{1}{c|}{23.5} & 55.2 & 16.1 & \textbf{8.5} & \multicolumn{1}{c|}{\textbf{12.5}} & 23.2 & 0.9 & \textbf{5.1} & \multicolumn{1}{c|}{3.4} & 25.9 \\
      & CLAP   & \textbf{89.6} & \textbf{52.2} & \textbf{37.1} & \multicolumn{1}{c|}{\textbf{23.9}} & \textbf{73.4} & \textbf{21.2} & 7.4 & \multicolumn{1}{c|}{\textbf{12.5}} & \textbf{27.0} & \textbf{1.1} & 5.0 & \multicolumn{1}{c|}{\textbf{3.5}} & \textbf{29.5} \\
    \bottomrule
    \end{tabular}
    }

\end{table}

\subsection{More Analysis}
\subsubsection{t-SNE Visualization} 
\begin{figure}[tb]
  \centering
  \resizebox{1\linewidth}{!}{
    \begin{subfigure}{0.32\linewidth}
    \includegraphics[width=\linewidth]{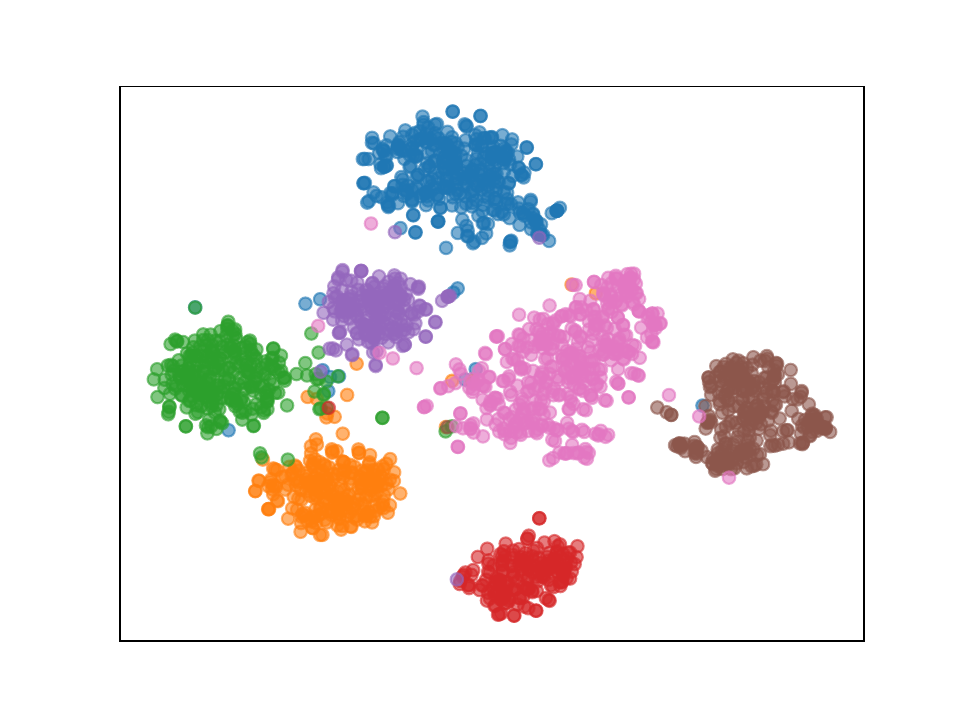}
    \caption{CLIP}
    \end{subfigure}
    
    \begin{subfigure}{0.32\linewidth}
    \includegraphics[width=\linewidth]{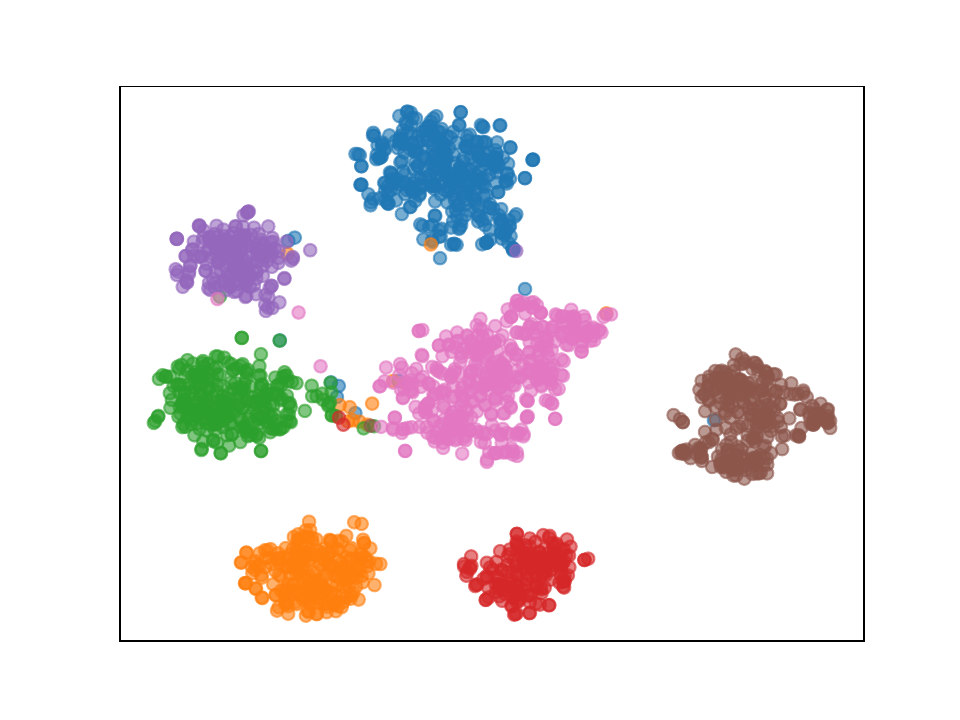}
    \caption{Im.Aug}
    \end{subfigure}
    
    \begin{subfigure}{0.32\linewidth}
    \includegraphics[width=\linewidth]{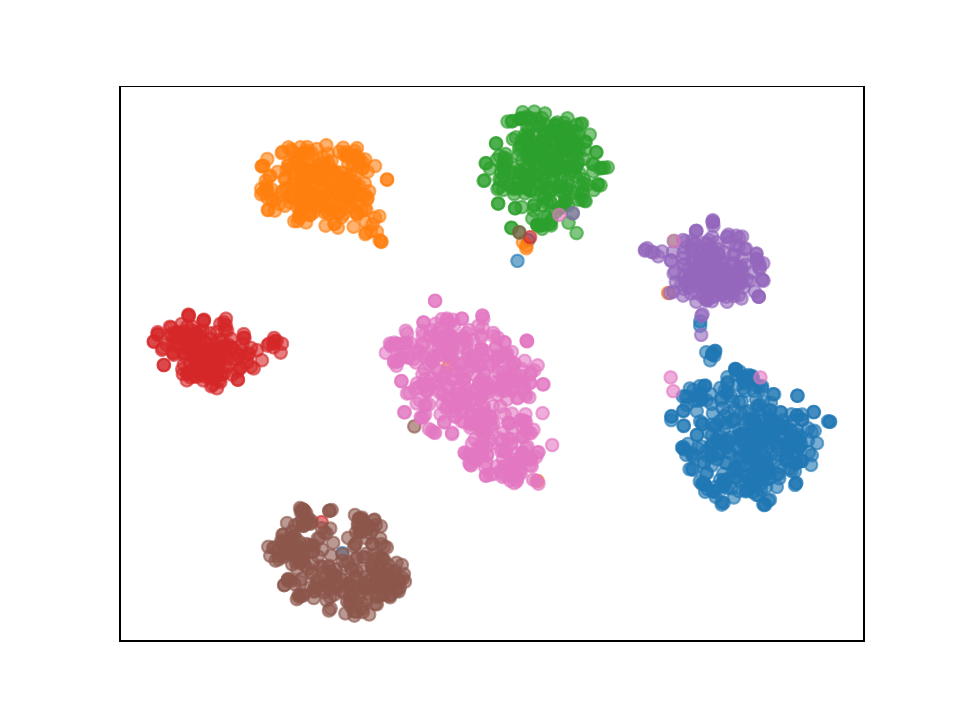}
    \caption{CLAP}
    \end{subfigure}
    }
  \caption{t-SNE visualizations of all images in the Art Painting of PACS dataset show CLAP outperforms the original CLIP and Im.Aug, with clearer inter-class distinctions and tighter intra-class clusters.}
  \label{fig:t-SNE}
  
\end{figure}

In our t-SNE visualizations, we examine the representations of CLIP, Im.Aug, and CLAP for all images within the Art Painting domain of the PACS dataset. \cref{fig:t-SNE} shows that CLAP's image representations display a marked inter-class separation and tighter intra-class clustering than those of CLIP and Im.Aug. This observation suggests that CLAP's representations are more closely tied to content information and less influenced by style information, in contrast to the other two.

\subsubsection{Ablations}
\begin{figure}[tb]
  \centering
  \resizebox{1\linewidth}{!}{
      \begin{subfigure}{0.3\linewidth}
          \includegraphics[width=\linewidth]{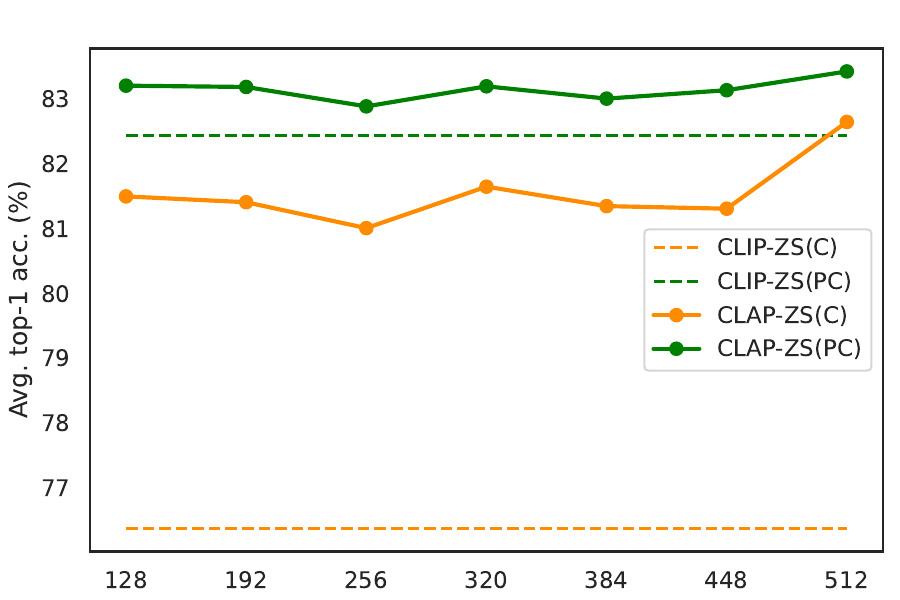}
          \caption{Latent dimensions}
          \label{subfig:latentdim}
      \end{subfigure}
      
      \begin{subfigure}{0.3\linewidth}
          \includegraphics[width=\linewidth]{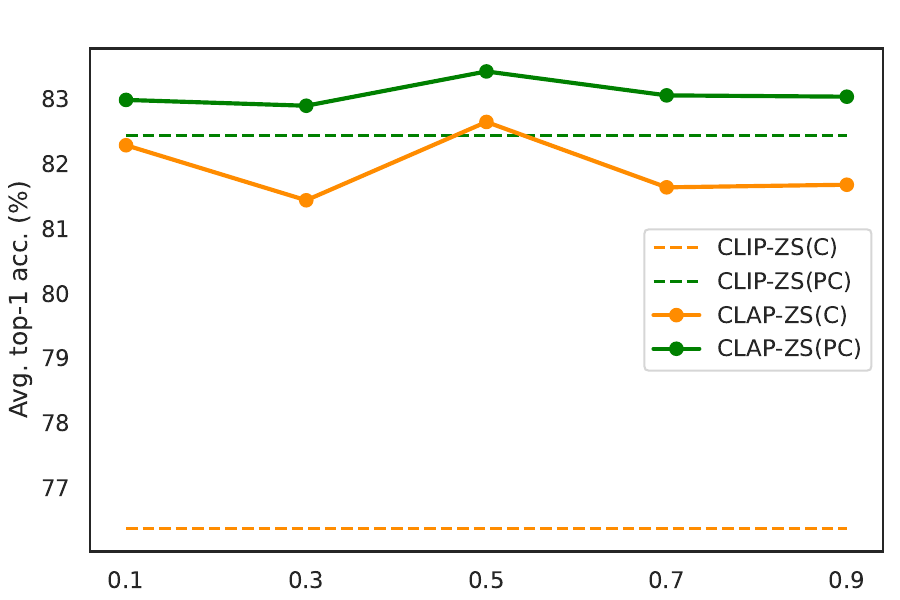}
          \caption{Temperature $\tau$ in loss}
          \label{subfig:tau}
      \end{subfigure}
      
      \begin{subfigure}{0.3\linewidth}
          \includegraphics[width=\linewidth]{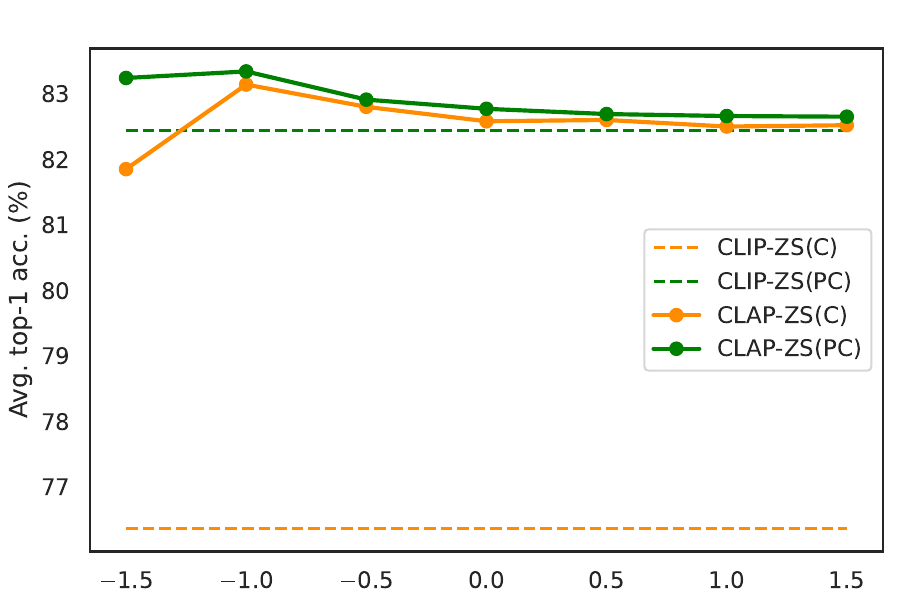}
          \caption{Inference weight $\mathrm{log}_{10}(\alpha)$}
          \label{subfig:alpha}
      \end{subfigure}
  }
  \caption{We conduct ablative study on hyper-parameter choices on the VLCS dataset, including latent dimensions, $\tau$ values, and $\alpha$ values during the inference stage. CLAP continuously enhance CLIP's performance throughout the tested values.}
  \label{fig:ablations}
\end{figure}

In \cref{fig:ablations}, we assess the zero-shot capabilities of our model using two distinct prompts, ZS(C) and ZS(PC), on the VLCS dataset. This analysis forms part of an ablative study aimed at understanding the influence of various hyper-parameters on model performance. Specifically, we examine: the dimensions of the latent layer within the MLP of the disentangled network, as illustrated in \cref{subfig:latentdim}; the temperature parameter ($\tau$) in the loss function, as depicted in \cref{subfig:tau}; and the weight coefficient ($\alpha$) during the inference stage, as shown in \cref{subfig:alpha}. Our findings indicate that CLAP consistently enhances zero-shot performance across all tested configurations for both prompts, while also significantly reducing the gap between the performances elicited by each prompt. These results underscore the efficacy of CLAP in accommodating a wide range of hyper-parameters.

\subsubsection{Prompt Augmentation Combinations} We explore diverse combinations of our tailored prompt augmentation methods and examine Easy Data Augmentation (EDA) techniques \cite{wei2019eda} on the VLCS dataset. Each tested technique showcases CLAP's enhancements over CLIP, with details available in \cref{subsec:augcombo}.

\subsubsection{Prompt Sources} We assess the impact of different training prompt formats, originating from various synthetic sources, on the performance of the VLCS dataset, incorporating EDA techniques. Our evaluation includes our template-based prompts, LLM-generated prompts by ChatGPT-3.5 \cite{brown2020language} (with the generation process detailed in \cref{subsec:llm-gen}
), prompts structured as "a [random] style of [class]," where "[random]" is filled with terms from a random word generator\footnote{\url{https://github.com/vaibhavsingh97/random-word}}, and prompts produced using the PromptStyler method \cite{cho2023promptstyler}. The findings indicate that the training prompts with simpler forms tend to yield better performance, with detailed quantitative results presented in \cref{subsec:promptsource}.

\subsubsection{Experiments on Different Model Scales} In our repeated experiments assessing zero-shot performance on the ViT-L/14 and ResNet50x16 pre-trained with CLIP, we consistently find that CLAP improves zero-shot performance while also reducing performance variances. This consistent observation underscores CLAP's effectiveness in enhancing the quality of CLIP representations. For quantitative details supporting these findings, please see the \cref{sec:repeated_exps}.

\section{Conclusion}
To enhance pre-trained CLIP-like models, this study delves into disentangling latent content variables. Through a causal analysis of the underlying generative processes of vision-language data, we discover that training a disentangled network in one modality can effectively disentangle content across both modalities. Given the high semantic nature of text data, we identify that disentanglement is more achievable within the language modality through text augmentation interventions. Building on these insights, we introduce CLAP (Contrastive Learning with Augmented Prompts) to acquire disentangled vision-language content features. Comprehensive experiments validate CLAP's effectiveness, demonstrating significant improvements in zero-shot and few-shot performance, and enhancing robustness against perturbations. We anticipate that our work will inspire further exploration into disentangling latent variables within vision-language models.

% \newpage
% \clearpage\mbox{}Page \thepage\ of the manuscript.
% \clearpage\mbox{}Page \thepage\ of the manuscript.
% \clearpage\mbox{}Page \thepage\ of the manuscript.
% \clearpage\mbox{}Page \thepage\ of the manuscript.
% \clearpage\mbox{}Page \thepage\ of the manuscript. This is the last page.
% Now we have reached the maximum length of an ECCV \ECCVyear{} submission (excluding references).
% References should start immediately after the main text, but can continue past p.\ 14 if needed.

% \clearpage  % TODO REVIEW/FINAL: This \clearpage needs to be removed from both review and camera-ready versions.


\bibliographystyle{splncs04}
\bibliography{main}

\appendix
\title{ CLAP: Isolating Content from Style through\\ Contrastive Learning with Augmented Prompts \\{\small APPENDIX}}
\author{Yichao Cai
\orcidlink{0000-0003-1607-8948}\textsuperscript{(\Letter)}
\and Yuhang Liu
\orcidlink{0000-0002-8195-9349} 
\and Zhen Zhang
\orcidlink{0000-0003-2805-4396} 
\and Javen Qinfeng Shi
\orcidlink{0000-0002-9126-2107}
}

    \titlerunning{APPENDIX}
\authorrunning{Y.~Cai et al.}

\institute{Australian Institute for Machine Learning, University of Adelaide, SA 5000, Australia \\
\email{\{yichao.cai,yuhang.liu01,zhen.zhang02,javen.shi\}@adelaide.edu.au}\\
}
\maketitle

\subsubsection{Overview of the Appendix:} 
\begin{itemize}
\item More details on experiments using the CLIP pre-trained ViT-B/16 model are provided in \cref{sec:vitb-exp}, including implementation details in \cref{subsec:implementdetail}, investigations into prompt augmentation combinations in \cref{subsec:augcombo}, analysis of different training prompt sources in \cref{subsec:promptsource}, and detailed experiment results for each dataset in \cref{subsec:vitb-details}.
\item The processes of data synthesis with large models used in our approach are outlined in \cref{sec:data-syn}: The image synthesis procedure for Im.Aug is detailed in \cref{subsec:sd-gen}, and the approach for generating "LLM" prompts, used in analyzing prompt sources, is described in \cref{subsec:llm-gen}.
\item In \cref{sec:repeated_exps}, we detail our repeated zero-shot experiments conducted with the CLIP pre-trained ViT-L/14 (\cref{subsec:vitl14}) and ResNet50x16 (\cref{subsec:rn50x16}) models.
\item In section \cref{sec:dicussison}, we present discussions covering the underlying rationale for basing CLAP on the CLIP pre-trained models in \cref{subsec:rationale}, and the impact of image augmentation and text augmentation in \cref{subsec:impact_augmentations}.
\end{itemize}

\section{More on Experiments with ViT-B/16} 
\label{sec:vitb-exp}

\subsection{Implementation Details}

\label{subsec:implementdetail}

In this section, we detail the implementation of our experiments utilizing the CLIP pre-trained ViT-B/16 model:

\paragraph{Network.} The network's output dimension is aligned with the 512-dimensional CLIP features, thereby obviating the need for input feature downsampling. The latent dimensions are tailored to each dataset: 256 for PACS, 448 for OfficeHome, and 512 for VLCS and DomainNet, to accommodate the variety of categories and complexity of datasets. The weight parameter $\alpha$ is adjusted to 0.208 for PACS, 0.056 for VLCS, 0.14 for OfficeHome, and 0.2 for DomainNet, while it is consistently maintained at 1 throughout the training phase.

\paragraph{Training CLAP.} Training parameters are consistent across datasets, employing the Adam optimizer with a learning rate of 0.0001, limiting training to 8,000 steps with checking the average loss every 480 steps, and instituting early stopping after five checkpoints without a loss decrease of at least 0.01. Batch sizes are adjusted to 8 for PACS and VLCS, 96 for OfficeHome, and 384 for DomainNet, with the temperature parameter $\tau$ set at 0.5 for PACS and VLCS, and 0.3 for OfficeHome and DomainNet. The loss coefficient $\lambda$ is set to 1 for PACS and VLCS, and 0.0001 for OfficeHome and DomainNet, due to the first two datasets have less classes. Prompt augmentations, OSD+OCD+SPO, are applied across datasets all with a 0.5 probability. For the PACS and VLCS datasets, Gaussian noise with a zero mean and a standard deviation of 0.02 is randomly inserted at the beginning, middle, or end of the augmented-view prompts to enrich the training samples. In the linear probe evaluations for few-shot analysis, L2 normalization and cross-entropy loss are utilized for training over 1,000 epochs with a batch size of 32, incorporating early stopping with a patience threshold of 10 epochs and a loss decrease criterion of 0.001.

\paragraph{Training Im.Aug.} We train a disentangled network using image augmentation, applying the InfoNCE loss with a temperature parameter $\tau$ set to 0.5. This include image augmentation techniques, image cropping ($scale \in [0.64, 1.0]$) and color distortion ($brightness=0.5, hue=0.3$), each with a probability of 0.5. Other training and inference configurations for Im.Aug are consistent with those used for CLAP across all datasets.

\subsection{Prompt Augmentation Combinations}
\label{subsec:augcombo}
In \cref{tab:augmentation}, we explore different combinations of our tailored prompt augmentation techniques and EDA (Easy Data Augmentation) \cite{wei2019eda} techniques on the VLCS dataset. Each combination demonstrates CLAP's effectiveness in enhancing CLIP's performance and reducing performance disparities. The combination of OSD+OCS+SPO+IGN achieves the highest average accuracy and the least variance, outperforming the EDA techniques. Notably, even without incorporating random noise in the augmentations, CLAP significantly surpasses CLIP in handling perturbations on prompts, as evidenced by the largely reduced $\Delta_{(NC)}$.

\begin{table}[tb]
\caption{We evaluate prompt augmentation combinations on the VLCS dataset: OSD (\textcircled{1}), OCD (\textcircled{2}), ITD (\textcircled{3}), ASD (\textcircled{4}), SPO (\textcircled{5}), and IGN (\textcircled{6}). ZS(Avg.) shows average zero-shot accuracy acoss four distinct inference prompts. CLAP boosts CLIP's accuracy and reduces variances, with \textcircled{1}\textcircled{2}\textcircled{5}\textcircled{6} as the optimal combination.}
\label{tab:augmentation}
\centering
    % \resizebox{\linewidth}{!}{
    \begin{tabular}{@{}ccccccccc@{}}
    \toprule
    \multirow{3}{*}{Metrics} & \multirow{3}{*}{\begin{tabular}[c]{@{}c@{}}CLIP\\      (base)\end{tabular}} & \multicolumn{7}{c}{Avg. top-1 acc. (\%)   of different augmentations} \\ \cmidrule(l){3-9} 
                             &                                                                                 & \multirow{2}{*}{EDA}   &\textcircled{1}\textcircled{2}\textcircled{3}   & \textcircled{1}\textcircled{2}\textcircled{3}    & \textcircled{1}\textcircled{2}\textcircled{3}   & \textcircled{1}\textcircled{2}\textcircled{3}   & \textcircled{3}\textcircled{4}\textcircled{5}   & \textcircled{1}\textcircled{2}\textcircled{5}           \\
                             &                                                                                 &                        & \textcircled{4}\textcircled{5}\textcircled{6}    & \textcircled{4}\textcircled{5}    & \textcircled{4}\textcircled{6}    & \textcircled{4}     & \textcircled{6}     & \textcircled{6}              \\ \midrule
    ZS(Avg.) (↑)             & 77.3                                                                            & 81.6                   & 82.0   & 80.1   & 82.0  & 79.6  & 82.1  & \textbf{82.6}  \\ \midrule
    $R$ (↓)                    & 6.1                                                                             & 1.9                    & 1.2    & 2.5    & 0.9   & 3.2   & 1.6   & \textbf{0.8}   \\
    $\delta$ (↓)                    & 2.8                                                                             & 0.9                    & 0.6    & 1.2    & \textbf{0.4}   & 1.5   & 0.7   & \textbf{0.4}   \\
    $\Delta_{(NC)}$   (↓)            & 8.1                                                                             & 2.3                    & 1.7    & 3.0    & 1.8   & 3.4   & 2.0   & \textbf{1.6}   \\ \bottomrule
    \end{tabular}
    % }
\end{table}

\subsection{Prompt Sources}  
\label{subsec:promptsource}

In \cref{tab:promptsource}, we examine the effects of various training prompt formats, sourced from different synthetic origins, on the VLCS dataset performance, utilizing
\begin{table}[tb]
\caption{We employ EDA augmentation to train CLAP with diverse prompt sources on the VLCS dataset. Each prompt source contributes to improvements in CLIP's zero-shot performance, with "Random" and "Template" prompts, in their simpler forms, yielding better outcomes.}
\label{tab:promptsource}
\centering
    % \resizebox{\linewidth}{!}{
    \begin{tabular}{@{}cccccc@{}}
    \toprule
    \multirow{2}{*}{Metrics} & \multirow{2}{*}{\begin{tabular}[c]{@{}c@{}}CLIP\\ (base)\end{tabular}} & \multicolumn{4}{c}{Avg. top-1 acc. (\%)   of different sources} \\ \cmidrule(l){3-6} 
                             &                                                                        & ~~LLM~~       & ~Random~             & Prm.Stl.      & Template          \\ \midrule
    ZS(Avg.) (↑)             & 77.3                                                                   & 78.2      & \textbf{81.6}      & 81.2              & \textbf{81.6}     \\ \midrule
    $R$ (↓)                    & 6.1                                                                    & 3.2       & \textbf{0.7}       & 2.7               & 1.9               \\
    $\delta$ (↓)                    & 2.8                                                                    & 1.5       & \textbf{0.3}       & 1.2               & 0.9               \\
    $\Delta_{(NC)}$   (↓)            & 8.1                                                                    & 3.3       & \textbf{2.3}       & 3.0               & \textbf{2.3}      \\ \bottomrule
    \end{tabular}
    % }
\end{table}

EDA techniques. The prompt formats are defined as follows: "Template" refers to the template-based prompts fundamental to our primary approach; "LLM" designates prompts created by ChatGPT-3.5 \cite{brown2020language}, with the generation process elaborated in \cref{subsec:llm-gen}; "Random" describes prompts formatted as "a [random] style of [class]," with "[random]" being replaced by terms from a random word generator; and "Prm.Stl." indicates vectorized prompts generated through PromptStyler\cite{cho2023promptstyler}.

Our experimental results indicate that CLAP, when trained across these varied prompt formats, enhances the performance of CLIP. Notably, despite the complex generation mechanisms of "LLM" and "Prm.Stl." prompts, the simpler, random-styled and template-based prompts demonstrate superior efficacy. However, it is important to highlight that the improvements attributed to these diverse prompt formats, trained with EDA, do not surpass the best performance of the prompt augmentations tailored for template-based prompts.

\subsection{Detailed Results on ViT-B/16}
\label{subsec:vitb-details}

\subsubsection{Details on Zero-Shot Evaluations}
\label{subsec:zero-shot}
We present the domain-level zero-shot performance with various prompts across each dataset in \cref{tab:zs-vitb-dm}. CLAP consistently enhances CLIP's zero-shot performance across these different prompts. Given that CLAP exclusively utilizes text data for training, it does not compromise CLIP's inherent ability to generalize across domains, which is acquired from its extensive training dataset. Rather, by achieving a more effective disentanglement of content, it unequivocally enhances CLIP's zero-shot performance across all dataset domains.

\begin{table}[tbp]
\caption{Domain-level zero-shot top-1 accuracy (\%) with the ViT-B/16 CLIP backbone. Panels (a) and (b) report results for the four inference prompt constructions.}
\label{tab:zs-vitb-dm}
\centering
\small
\setlength{\tabcolsep}{4pt}
\renewcommand{\arraystretch}{1.05}

% ================================================================
% (a) ZS(C) and ZS(CP)
% ================================================================
\begin{tabular*}{\linewidth}{@{\extracolsep{\fill}}ll*{6}{c}@{}}
\toprule
\multicolumn{8}{c}{\textbf{(a) ZS(C) and ZS(CP)}} \\
\midrule
\multirow{2}{*}{Dataset}
& \multirow{2}{*}{Domain}
& \multicolumn{3}{c}{ZS(C)}
& \multicolumn{3}{c}{ZS(CP)} \\
\cmidrule(lr){3-5}
\cmidrule(l){6-8}
&
& CLIP & Im.Aug & CLAP
& CLIP & Im.Aug & CLAP \\
\midrule

\multirow{4}{*}{PACS}
& A. & 96.4 & 96.9 & 97.5 & 93.4 & 97.0 & 97.6 \\
& C. & 98.9 & 99.0 & 98.9 & 99.0 & 99.2 & 99.0 \\
& P. & 99.9 & 99.9 & 99.9 & 99.3 & 99.6 & 99.9 \\
& S. & 87.7 & 90.1 & 92.5 & 89.2 & 89.6 & 92.5 \\
\midrule

\multirow{4}{*}{VLCS}
& C. & 99.7 & 99.8 & 99.9 & 99.9 & 99.9 & 99.9 \\
& L. & 61.8 & 66.2 & 67.7 & 69.9 & 70.4 & 70.4 \\
& S. & 70.1 & 74.8 & 78.0 & 73.3 & 76.0 & 77.2 \\
& V. & 73.9 & 77.1 & 84.9 & 84.8 & 85.4 & 86.0 \\
\midrule

\multirow{4}{*}{OfficeHome}
& A. & 80.5 & 79.0 & 81.8 & 80.1 & 76.0 & 81.6 \\
& C. & 64.6 & 59.6 & 66.4 & 63.7 & 58.9 & 65.4 \\
& P. & 86.3 & 83.6 & 87.5 & 86.6 & 83.4 & 87.2 \\
& R. & 88.0 & 85.9 & 88.5 & 87.6 & 84.8 & 87.7 \\
\midrule

\multirow{6}{*}{DomainNet}
& C. & 71.0 & 64.3 & 71.9 & 70.5 & 62.1 & 72.0 \\
& I. & 48.6 & 40.5 & 50.6 & 47.7 & 40.7 & 49.5 \\
& P. & 66.6 & 59.1 & 67.7 & 66.0 & 59.0 & 67.3 \\
& Q. & 14.9 & 12.4 & 15.2 & 13.3 & 11.5 & 13.8 \\
& R. & 82.6 & 76.6 & 83.1 & 82.2 & 75.8 & 82.2 \\
& S. & 63.1 & 56.1 & 63.7 & 62.2 & 55.0 & 63.1 \\
\bottomrule
\end{tabular*}

\vspace{1em}

% ================================================================
% (b) ZS(PC) and ZS(NC)
% ================================================================
\begin{tabular*}{\linewidth}{@{\extracolsep{\fill}}ll*{6}{c}@{}}
\toprule
\multicolumn{8}{c}{\textbf{(b) ZS(PC) and ZS(NC)}} \\
\midrule
\multirow{2}{*}{Dataset}
& \multirow{2}{*}{Domain}
& \multicolumn{3}{c}{ZS(PC)}
& \multicolumn{3}{c}{ZS(NC)} \\
\cmidrule(lr){3-5}
\cmidrule(l){6-8}
&
& CLIP & Im.Aug & CLAP
& CLIP & Im.Aug & CLAP \\
\midrule

\multirow{4}{*}{PACS}
& A. & 97.4 & 97.6 & 97.6 & 87.8 & 93.5 & 97.1 \\
& C. & 99.1 & 99.0 & 98.9 & 95.4 & 97.6 & 98.8 \\
& P. & 99.9 & 99.9 & 99.9 & 93.1 & 99.0 & 99.9 \\
& S. & 88.1 & 89.4 & 92.3 & 87.1 & 89.3 & 93.1 \\
\midrule

\multirow{4}{*}{VLCS}
& C. & 99.9 & 99.9 & 99.9 & 87.0 & 96.0 & 99.9 \\
& L. & 70.2 & 70.2 & 70.7 & 55.9 & 59.9 & 65.9 \\
& S. & 73.6 & 76.4 & 76.9 & 61.4 & 66.2 & 75.3 \\
& V. & 86.1 & 85.6 & 86.2 & 68.9 & 70.3 & 82.9 \\
\midrule

\multirow{4}{*}{OfficeHome}
& A. & 83.2 & 78.7 & 83.2 & 73.0 & 69.2 & 73.6 \\
& C. & 68.1 & 61.9 & 69.0 & 57.0 & 52.0 & 60.4 \\
& P. & 89.1 & 86.6 & 89.7 & 77.2 & 72.3 & 78.9 \\
& R. & 89.8 & 87.2 & 90.0 & 79.0 & 76.5 & 81.1 \\
\midrule

\multirow{6}{*}{DomainNet}
& C. & 71.3 & 63.4 & 72.8 & 63.2 & 53.9 & 64.6 \\
& I. & 47.8 & 40.0 & 50.5 & 42.9 & 35.0 & 45.1 \\
& P. & 66.5 & 59.8 & 68.4 & 57.2 & 50.4 & 59.4 \\
& Q. & 14.1 & 11.8 & 14.3 & 12.0 & 9.2 & 13.1 \\
& R. & 83.4 & 78.2 & 83.7 & 75.2 & 67.9 & 75.6 \\
& S. & 63.4 & 56.4 & 64.4 & 55.7 & 47.5 & 57.6 \\
\bottomrule
\end{tabular*}

\end{table}

\begin{longtable}{@{}lllccccc@{}}
\caption{Domain-level few-shot linear-probe top-1 accuracy (\%) with the ViT-B/16 CLIP backbone.}
\label{tab:fs-vitb-dm}\\

\toprule
Dataset & Domain & Method
& 1-shot & 4-shot & 8-shot & 16-shot & 32-shot \\
\midrule
\endfirsthead

\multicolumn{8}{c}{\tablename\ \thetable\ --- continued from previous page} \\
\toprule
Dataset & Domain & Method
& 1-shot & 4-shot & 8-shot & 16-shot & 32-shot \\
\midrule
\endhead

\midrule
\multicolumn{8}{r}{Continued on next page} \\
\endfoot

\bottomrule
\endlastfoot

% ================================================================
% PACS
% ================================================================
\multirow{12}{*}{PACS}
& \multirow{3}{*}{A.}
& CLIP   & 79.5 & 92.4 & 95.1 & 97.9 & 98.8 \\
& & Im.Aug & 84.1 & 96.4 & 97.2 & 98.1 & 99.1 \\
& & CLAP   & 94.5 & 97.2 & 98.4 & 98.4 & 98.9 \\
\cmidrule(l){2-8}

& \multirow{3}{*}{C.}
& CLIP   & 86.7 & 96.8 & 98.8 & 99.5 & 99.6 \\
& & Im.Aug & 96.1 & 98.6 & 98.9 & 99.2 & 99.6 \\
& & CLAP   & 98.3 & 99.2 & 99.3 & 99.5 & 99.6 \\
\cmidrule(l){2-8}

& \multirow{3}{*}{P.}
& CLIP   & 97.4 & 99.6 & 99.9 & 99.8 & 99.9 \\
& & Im.Aug & 99.8 & 99.8 & 99.9 & 99.9 & 99.9 \\
& & CLAP   & 99.9 & 99.9 & 99.9 & 99.9 & 99.9 \\
\cmidrule(l){2-8}

& \multirow{3}{*}{S.}
& CLIP   & 75.1 & 91.1 & 92.3 & 92.4 & 93.9 \\
& & Im.Aug & 80.0 & 92.3 & 92.3 & 92.6 & 94.2 \\
& & CLAP   & 87.3 & 92.5 & 92.9 & 93.1 & 94.1 \\
\midrule

% ================================================================
% VLCS
% ================================================================
\multirow{12}{*}{VLCS}
& \multirow{3}{*}{C.}
& CLIP   & 99.2 & 99.9 & 99.8 & 99.7 & 99.9 \\
& & Im.Aug & 99.7 & 99.8 & 99.7 & 99.9 & 100.0 \\
& & CLAP   & 99.8 & 99.9 & 99.9 & 99.9 & 99.9 \\
\cmidrule(l){2-8}

& \multirow{3}{*}{L.}
& CLIP   & 41.3 & 56.7 & 46.2 & 59.4 & 60.4 \\
& & Im.Aug & 41.3 & 57.0 & 36.8 & 60.4 & 60.7 \\
& & CLAP   & 41.1 & 59.8 & 48.3 & 62.6 & 61.9 \\
\cmidrule(l){2-8}

& \multirow{3}{*}{S.}
& CLIP   & 45.3 & 61.9 & 67.4 & 75.9 & 77.4 \\
& & Im.Aug & 46.1 & 63.7 & 67.7 & 76.8 & 78.6 \\
& & CLAP   & 50.8 & 69.0 & 71.3 & 80.9 & 81.0 \\
\cmidrule(l){2-8}

& \multirow{3}{*}{V.}
& CLIP   & 50.9 & 64.5 & 75.4 & 72.6 & 85.7 \\
& & Im.Aug & 53.4 & 66.7 & 74.1 & 73.9 & 86.1 \\
& & CLAP   & 59.0 & 76.1 & 78.7 & 77.7 & 87.9 \\
\midrule

% ================================================================
% OfficeHome
% ================================================================
\multirow{12}{*}{OfficeHome}
& \multirow{3}{*}{A.}
& CLIP   & 42.6 & 76.8 & 84.8 & 91.8 & 97.4 \\
& & Im.Aug & 45.1 & 77.6 & 86.0 & 92.1 & 97.5 \\
& & CLAP   & 43.9 & 77.7 & 85.5 & 92.1 & 97.5 \\
\cmidrule(l){2-8}

& \multirow{3}{*}{C.}
& CLIP   & 40.1 & 69.9 & 75.8 & 81.6 & 89.0 \\
& & Im.Aug & 45.0 & 70.2 & 75.9 & 81.6 & 89.0 \\
& & CLAP   & 43.8 & 70.5 & 76.6 & 81.6 & 89.2 \\
\cmidrule(l){2-8}

& \multirow{3}{*}{P.}
& CLIP   & 70.2 & 89.7 & 93.8 & 95.7 & 97.7 \\
& & Im.Aug & 73.3 & 90.3 & 93.7 & 95.7 & 97.6 \\
& & CLAP   & 73.4 & 90.2 & 93.9 & 95.8 & 97.6 \\
\cmidrule(l){2-8}

& \multirow{3}{*}{R.}
& CLIP   & 58.4 & 81.7 & 89.7 & 92.9 & 95.8 \\
& & Im.Aug & 59.3 & 83.1 & 89.5 & 92.7 & 95.8 \\
& & CLAP   & 59.4 & 82.9 & 89.9 & 93.2 & 95.8 \\
\midrule

% ================================================================
% DomainNet
% ================================================================
\multirow{18}{*}{DomainNet}
& \multirow{3}{*}{C.}
& CLIP   & 42.1 & 66.8 & 74.2 & 78.5 & 82.8 \\
& & Im.Aug & 43.6 & 67.5 & 74.3 & 78.6 & 82.8 \\
& & CLAP   & 43.8 & 67.8 & 74.6 & 78.8 & 82.7 \\
\cmidrule(l){2-8}

& \multirow{3}{*}{I.}
& CLIP   & 19.5 & 38.5 & 46.7 & 53.2 & 60.0 \\
& & Im.Aug & 20.8 & 39.3 & 47.0 & 53.2 & 59.9 \\
& & CLAP   & 21.0 & 39.7 & 47.3 & 53.6 & 60.1 \\
\cmidrule(l){2-8}

& \multirow{3}{*}{P.}
& CLIP   & 32.1 & 60.5 & 68.0 & 72.5 & 76.7 \\
& & Im.Aug & 33.5 & 60.9 & 68.0 & 72.6 & 76.6 \\
& & CLAP   & 34.2 & 61.5 & 68.7 & 73.0 & 76.8 \\
\cmidrule(l){2-8}

& \multirow{3}{*}{Q.}
& CLIP   & 15.2 & 30.0 & 37.1 & 43.8 & 49.4 \\
& & Im.Aug & 15.3 & 29.6 & 36.4 & 43.4 & 49.1 \\
& & CLAP   & 15.3 & 29.9 & 36.8 & 43.5 & 49.0 \\
\cmidrule(l){2-8}

& \multirow{3}{*}{R.}
& CLIP   & 50.8 & 76.7 & 81.7 & 84.0 & 85.9 \\
& & Im.Aug & 52.1 & 77.0 & 81.9 & 83.9 & 85.9 \\
& & CLAP   & 52.7 & 77.6 & 82.2 & 84.3 & 86.0 \\
\cmidrule(l){2-8}

& \multirow{3}{*}{S.}
& CLIP   & 33.1 & 56.2 & 62.9 & 67.8 & 72.5 \\
& & Im.Aug & 33.9 & 56.6 & 62.9 & 67.7 & 72.3 \\
& & CLAP   & 34.8 & 57.2 & 63.7 & 68.1 & 72.6 \\

\end{longtable}

\subsubsection{Details on Few-Shot Evaluations}
\label{subsec:few-shot}
We display the quantitative results of few-shot performance in \cref{tab:fs-vitb-dm}. CLAP consistently enhances the few-shot capabilities, showcasing improvements across test datasets at a closer domain level.

\subsubsection{Details on Adversarial Evaluations}
\label{subsec:adversarial}

In \cref{tab:adv-vitb-dm}, we detail our adversarial performance evaluations for PACS, VLCS, OfficeHome, and DomainNet, respectively. CLAP generally improves both zero-shot and one-shot adversarial robustness, although the magnitude and direction of the gains vary across individual domains and attacks. While Im.Aug boosts one-shot robustness against adversarial tasks, its impact on zero-shot adversarial robustness is inconsistent.

\subsubsection{Details on Ablative Analysis}
\label{subsec-abla}
In \cref{tab:abla-pmp}, we provide detailed results from our analysis on zero-shot performance using various combinations of prompt augmentations. Additionally, in \cref{tab:ablative-dm}, we present the outcomes of our ablative studies focusing on the hyperparameters $\tau$, latent dimension, and $\alpha$, respectively, each evaluated domain-wise. The results indicate that CLAP is effective across a wide range of hyperparameters.

\begingroup
\small
\setlength{\tabcolsep}{4.5pt}
\renewcommand{\arraystretch}{1.05}

\begin{longtable}{@{}lllcccccc@{}}
\caption{Domain-level top-1 accuracy (\%) under FGSM, PGD-20, and CW-20 adversarial attacks with the ViT-B/16 CLIP backbone.}
\label{tab:adv-vitb-dm}\\

\toprule
\multirow{2}{*}{Dataset}
& \multirow{2}{*}{Domain}
& \multirow{2}{*}{Attack}
& \multicolumn{3}{c}{ZS(C)}
& \multicolumn{3}{c}{1-shot} \\
\cmidrule(lr){4-6}
\cmidrule(l){7-9}
& & & CLIP & Im.Aug & CLAP & CLIP & Im.Aug & CLAP \\
\midrule
\endfirsthead

\multicolumn{9}{c}{\tablename\ \thetable\ --- continued from previous page} \\
\toprule
\multirow{2}{*}{Dataset}
& \multirow{2}{*}{Domain}
& \multirow{2}{*}{Attack}
& \multicolumn{3}{c}{ZS(C)}
& \multicolumn{3}{c}{1-shot} \\
\cmidrule(lr){4-6}
\cmidrule(l){7-9}
& & & CLIP & Im.Aug & CLAP & CLIP & Im.Aug & CLAP \\
\midrule
\endhead

\midrule
\multicolumn{9}{r}{Continued on next page} \\
\endfoot

\bottomrule
\endlastfoot

% ================================================================
% PACS
% ================================================================
\multirow{12}{*}{PACS}
& \multirow{3}{*}{A.}
& FGSM   & 76.3 & 79.3 & 79.3 & 61.2 & 78.0 & 87.3 \\
& & PGD-20 & 1.7 & 2.2 & 1.8 & 16.0 & 42.1 & 63.1 \\
& & CW-20  & 1.5 & 2.0 & 2.3 & 0.5 & 1.1 & 1.7 \\
\cmidrule(l){2-9}

& \multirow{3}{*}{C.}
& FGSM   & 94.9 & 95.0 & 94.0 & 66.5 & 84.2 & 95.1 \\
& & PGD-20 & 33.3 & 37.7 & 35.6 & 33.3 & 57.2 & 86.1 \\
& & CW-20  & 28.8 & 34.0 & 33.2 & 11.9 & 23.6 & 31.8 \\
\cmidrule(l){2-9}

& \multirow{3}{*}{P.}
& FGSM   & 91.6 & 90.3 & 91.7 & 67.4 & 80.8 & 92.1 \\
& & PGD-20 & 5.7 & 7.0 & 6.7 & 27.1 & 55.0 & 69.8 \\
& & CW-20  & 4.7 & 4.9 & 5.8 & 0.7 & 2.7 & 4.1 \\
\cmidrule(l){2-9}

& \multirow{3}{*}{S.}
& FGSM   & 84.5 & 87.5 & 89.8 & 71.6 & 74.6 & 83.8 \\
& & PGD-20 & 75.8 & 78.4 & 79.2 & 63.0 & 66.3 & 74.6 \\
& & CW-20  & 74.5 & 76.8 & 77.9 & 62.7 & 65.4 & 70.3 \\
\midrule

% ================================================================
% VLCS
% ================================================================
\multirow{12}{*}{VLCS}
& \multirow{3}{*}{C.}
& FGSM   & 98.4 & 98.4 & 98.1 & 77.2 & 78.6 & 93.4 \\
& & PGD-20 & 5.4 & 5.1 & 8.3 & 38.2 & 40.6 & 54.3 \\
& & CW-20  & 4.4 & 4.8 & 5.8 & 2.3 & 2.8 & 3.6 \\
\cmidrule(l){2-9}

& \multirow{3}{*}{L.}
& FGSM   & 47.0 & 54.0 & 56.0 & 37.8 & 37.1 & 39.3 \\
& & PGD-20 & 0.3 & 0.4 & 0.7 & 21.4 & 18.0 & 22.7 \\
& & CW-20  & 0.2 & 0.3 & 0.7 & 0.6 & 0.7 & 0.7 \\
\cmidrule(l){2-9}

& \multirow{3}{*}{S.}
& FGSM   & 53.1 & 58.8 & 60.6 & 25.9 & 27.9 & 28.7 \\
& & PGD-20 & 0.2 & 0.2 & 0.5 & 2.8 & 3.9 & 3.8 \\
& & CW-20  & 0.1 & 0.2 & 0.3 & 0.0 & 0.0 & 0.0 \\
\cmidrule(l){2-9}

& \multirow{3}{*}{V.}
& FGSM   & 64.0 & 67.1 & 72.9 & 39.9 & 42.9 & 47.4 \\
& & PGD-20 & 1.9 & 2.9 & 3.5 & 1.5 & 1.8 & 4.1 \\
& & CW-20  & 1.2 & 1.7 & 2.6 & 0.0 & 0.1 & 0.1 \\
\midrule

% ================================================================
% OfficeHome
% ================================================================
\multirow{12}{*}{OfficeHome}
& \multirow{3}{*}{A.}
& FGSM   & 55.3 & 53.8 & 55.5 & 25.8 & 28.8 & 25.3 \\
& & PGD-20 & 4.4 & 5.1 & 4.7 & 2.0 & 5.2 & 2.5 \\
& & CW-20  & 2.9 & 3.1 & 3.5 & 0.7 & 1.2 & 1.0 \\
\cmidrule(l){2-9}

& \multirow{3}{*}{C.}
& FGSM   & 49.4 & 45.5 & 50.6 & 27.0 & 32.6 & 30.4 \\
& & PGD-20 & 15.2 & 14.9 & 16.0 & 6.4 & 8.9 & 8.0 \\
& & CW-20  & 12.4 & 11.2 & 13.0 & 6.1 & 8.3 & 7.7 \\
\cmidrule(l){2-9}

& \multirow{3}{*}{P.}
& FGSM   & 61.7 & 58.1 & 62.5 & 48.0 & 46.9 & 51.6 \\
& & PGD-20 & 13.2 & 13.9 & 14.0 & 8.6 & 10.7 & 10.0 \\
& & CW-20  & 9.2 & 8.8 & 10.2 & 8.3 & 7.9 & 8.4 \\
\cmidrule(l){2-9}

& \multirow{3}{*}{R.}
& FGSM   & 65.3 & 63.2 & 65.6 & 36.5 & 40.1 & 41.0 \\
& & PGD-20 & 7.5 & 7.9 & 7.9 & 5.3 & 9.4 & 8.9 \\
& & CW-20  & 5.2 & 4.8 & 5.6 & 2.9 & 2.8 & 2.9 \\
\midrule

% ================================================================
% DomainNet
% ================================================================
\multirow{18}{*}{DomainNet}
& \multirow{3}{*}{C.}
& FGSM   & 57.8 & 50.9 & 58.8 & 33.3 & 34.3 & 35.0 \\
& & PGD-20 & 21.6 & 18.7 & 22.8 & 18.4 & 19.6 & 20.0 \\
& & CW-20  & 15.8 & 12.5 & 16.6 & 7.0 & 7.5 & 7.8 \\
\cmidrule(l){2-9}

& \multirow{3}{*}{I.}
& FGSM   & 35.8 & 28.0 & 37.0 & 12.2 & 13.3 & 13.2 \\
& & PGD-20 & 6.1 & 3.7 & 6.7 & 4.6 & 5.3 & 5.1 \\
& & CW-20  & 3.3 & 1.9 & 3.7 & 0.9 & 0.9 & 0.9 \\
\cmidrule(l){2-9}

& \multirow{3}{*}{P.}
& FGSM   & 43.9 & 39.0 & 44.3 & 18.4 & 20.6 & 20.3 \\
& & PGD-20 & 3.1 & 2.8 & 3.3 & 8.6 & 10.4 & 9.9 \\
& & CW-20  & 1.8 & 1.3 & 1.9 & 0.3 & 0.3 & 0.3 \\
\cmidrule(l){2-9}

& \multirow{3}{*}{Q.}
& FGSM   & 12.9 & 10.3 & 13.2 & 10.9 & 10.8 & 11.1 \\
& & PGD-20 & 8.4 & 6.8 & 8.6 & 5.4 & 5.4 & 5.6 \\
& & CW-20  & 7.1 & 5.4 & 7.4 & 4.9 & 4.8 & 5.1 \\
\cmidrule(l){2-9}

& \multirow{3}{*}{R.}
& FGSM   & 62.1 & 55.9 & 62.4 & 34.5 & 35.9 & 36.5 \\
& & PGD-20 & 7.1 & 6.5 & 7.5 & 17.6 & 19.7 & 19.6 \\
& & CW-20  & 4.5 & 3.4 & 4.7 & 1.2 & 1.4 & 1.4 \\
\cmidrule(l){2-9}

& \multirow{3}{*}{S.}
& FGSM   & 49.1 & 43.3 & 49.7 & 25.7 & 26.0 & 27.5 \\
& & PGD-20 & 17.8 & 15.5 & 18.6 & 13.6 & 14.4 & 15.1 \\
& & CW-20  & 13.4 & 10.2 & 13.9 & 5.0 & 5.2 & 5.6 \\

\end{longtable}

\endgroup

\section{Data Synthesis}
\label{sec:data-syn}

\subsection{Synthetic Image Generation}
\label{subsec:sd-gen}

\begin{figure}[tb]
    \centering
    \resizebox{0.95\linewidth}{!}{
    \includegraphics{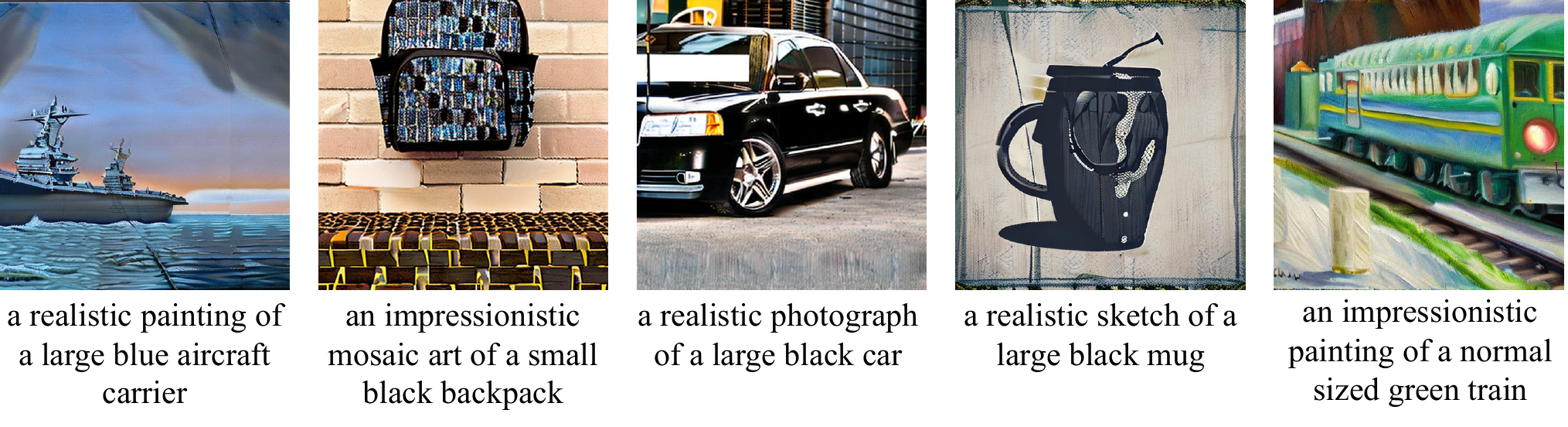}
    }
    \caption{Examples of synthetic images created with SDv2.1 and associated prompts.}
    \label{fig:syn-image}
\end{figure}

We employ the stable diffusion \cite{rombach2022high} v2.1 model for generating synthetic images used in our comparing experiments, specifically utilizing the Stable Diffusion v2-1 Model Card available on Hugging Face\footnote{\url{https://huggingface.co/stabilityai/stable-diffusion-2-1}}. For each class across the four datasets, we produce 480 images using our synthetic template prompts as input for the stable diffusion model. All generated images are of $512\times512$ resolution. Examples of these synthetic images alongside their corresponding text prompts are displayed in \cref{fig:syn-image}.

\begin{table}[tbp]
\caption{Domain-level zero-shot results on VLCS for different prompt-augmentation combinations and prompt sources. \textcircled{1}--\textcircled{6} denote OSD, OCD, ITD, ASD, SPO, and IGN, respectively. Panel (a) compares different prompt-augmentation policies, while Panel (b) compares different prompt sources using EDA.}
\label{tab:abla-pmp}
\centering
\small
\setlength{\tabcolsep}{3pt}
\renewcommand{\arraystretch}{1.08}

% ----------------------------------------------------------------
% Panel (a): Prompt-augmentation combinations
% ----------------------------------------------------------------
\begin{tabular*}{\linewidth}{@{\extracolsep{\fill}}cc*{8}{c}@{}}
\toprule
\multicolumn{10}{c}{\textbf{(a) Prompt-augmentation combinations}} \\
\midrule
\multirow{2}{*}{Prompt}
& \multirow{2}{*}{Domain}
& \multirow{2}{*}{CLIP}
& \multirow{2}{*}{EDA}
& \shortstack{\textcircled{1}\textcircled{2}\textcircled{3}\\\textcircled{4}\textcircled{5}\textcircled{6}}
& \shortstack{\textcircled{1}\textcircled{2}\textcircled{3}\\\textcircled{4}\textcircled{5}}
& \shortstack{\textcircled{1}\textcircled{2}\textcircled{3}\\\textcircled{4}\textcircled{6}}
& \shortstack{\textcircled{1}\textcircled{2}\textcircled{3}\\\textcircled{4}}
& \shortstack{\textcircled{3}\textcircled{4}\textcircled{5}\\\textcircled{6}}
& \shortstack{\textcircled{1}\textcircled{2}\textcircled{5}\\\textcircled{6}} \\
& & \multicolumn{8}{c}{Top-1 accuracy (\%) ($\uparrow$)} \\
\midrule

\multirow{4}{*}{ZS(C)}
& C. & 99.7 & 97.9 & 99.9 & 99.8 & 99.9 & 99.8 & 99.9 & 99.9 \\
& L. & 61.8 & 66.2 & 66.6 & 62.3 & 67.0 & 62.2 & 66.2 & 67.7 \\
& S. & 70.1 & 73.2 & 78.1 & 75.5 & 78.0 & 74.3 & 78.5 & 78.0 \\
& V. & 73.9 & 72.6 & 82.8 & 80.6 & 83.2 & 79.3 & 82.7 & 84.9 \\
\midrule

\multirow{4}{*}{ZS(CP)}
& C. & 99.9 & 99.8 & 99.9 & 99.9 & 99.9 & 99.9 & 99.9 & 99.9 \\
& L. & 69.9 & 69.3 & 69.3 & 67.9 & 69.6 & 68.4 & 70.0 & 70.4 \\
& S. & 73.3 & 76.2 & 77.6 & 76.4 & 76.7 & 75.9 & 78.8 & 77.2 \\
& V. & 84.8 & 77.0 & 85.3 & 84.0 & 85.3 & 84.2 & 85.1 & 86.0 \\
\midrule

\multirow{4}{*}{ZS(PC)}
& C. & 99.9 & 99.9 & 99.9 & 99.9 & 99.9 & 99.9 & 99.9 & 99.9 \\
& L. & 70.2 & 67.5 & 70.0 & 68.0 & 70.1 & 68.5 & 70.0 & 70.7 \\
& S. & 73.6 & 76.9 & 76.6 & 75.6 & 76.0 & 74.8 & 77.8 & 76.9 \\
& V. & 86.1 & 78.2 & 85.7 & 84.7 & 85.7 & 84.5 & 85.5 & 86.2 \\
\midrule

\multirow{4}{*}{ZS(NC)}
& C. & 87.0 & 95.3 & 99.8 & 99.6 & 99.8 & 99.4 & 99.7 & 99.9 \\
& L. & 55.9 & 63.0 & 65.2 & 61.3 & 65.6 & 60.5 & 65.4 & 65.9 \\
& S. & 61.4 & 68.9 & 75.6 & 70.3 & 75.2 & 68.3 & 74.9 & 75.3 \\
& V. & 68.9 & 69.3 & 80.1 & 75.2 & 80.4 & 73.8 & 79.4 & 82.9 \\
\bottomrule
\end{tabular*}

\vspace{1em}

% ----------------------------------------------------------------
% Panel (b): Prompt sources
% ----------------------------------------------------------------
\begin{tabular*}{\linewidth}{@{\extracolsep{\fill}}cc*{5}{c}@{}}
\toprule
\multicolumn{7}{c}{\textbf{(b) Prompt sources with EDA}} \\
\midrule
Prompt & Domain & CLIP & LLM & Random & Prm.Stl. & Template \\
\cmidrule(l){3-7}
& & \multicolumn{5}{c}{Top-1 accuracy (\%) ($\uparrow$)} \\
\midrule

\multirow{4}{*}{ZS(C)}
& C. & 99.7 & 97.9 & 99.7 & 99.9 & 99.9 \\
& L. & 61.8 & 66.2 & 69.0 & 67.3 & 66.5 \\
& S. & 70.1 & 73.2 & 76.9 & 73.5 & 76.9 \\
& V. & 73.9 & 72.6 & 81.8 & 81.8 & 81.9 \\
\midrule

\multirow{4}{*}{ZS(CP)}
& C. & 99.9 & 99.8 & 99.9 & 99.9 & 99.9 \\
& L. & 69.9 & 69.3 & 70.4 & 71.2 & 69.7 \\
& S. & 73.3 & 76.2 & 75.2 & 75.1 & 78.0 \\
& V. & 84.8 & 77.0 & 84.2 & 86.0 & 84.6 \\
\midrule

\multirow{4}{*}{ZS(PC)}
& C. & 99.9 & 99.9 & 99.9 & 99.9 & 99.9 \\
& L. & 70.2 & 67.5 & 70.6 & 71.8 & 70.0 \\
& S. & 73.6 & 76.9 & 75.1 & 74.9 & 78.2 \\
& V. & 86.1 & 78.2 & 84.6 & 86.8 & 84.8 \\
\midrule

\multirow{4}{*}{ZS(NC)}
& C. & 87.0 & 95.3 & 98.6 & 99.6 & 99.8 \\
& L. & 55.9 & 63.0 & 66.7 & 64.0 & 64.7 \\
& S. & 61.4 & 68.9 & 73.3 & 69.8 & 73.0 \\
& V. & 68.9 & 69.3 & 79.6 & 77.2 & 78.6 \\
\bottomrule
\end{tabular*}

\end{table}

\begin{table}[tbp]
\caption{Domain-level hyperparameter ablations on VLCS using the ViT-B/16 CLIP backbone. Each panel varies one hyperparameter while keeping the remaining configuration fixed. Values report zero-shot top-1 accuracy (\%).}
\label{tab:ablative-dm}
\centering
\small
\setlength{\tabcolsep}{5pt}
\renewcommand{\arraystretch}{1.05}

% ================================================================
% (a) Temperature
% ================================================================
\begin{tabular*}{\linewidth}{@{\extracolsep{\fill}}cc*{5}{c}@{}}
\toprule
\multicolumn{7}{c}{\textbf{(a) Temperature \(\tau\)}} \\
\midrule
Prompt & Domain
& \(\tau=0.1\)
& \(\tau=0.3\)
& \(\tau=0.5\)
& \(\tau=0.7\)
& \(\tau=0.9\) \\
\midrule
\multirow{4}{*}{ZS(C)}
& C. & 99.9 & 99.9 & 99.9 & 99.9 & 99.9 \\
& L. & 67.6 & 66.3 & 67.7 & 65.9 & 66.0 \\
& S. & 77.5 & 77.2 & 78.0 & 77.7 & 77.6 \\
& V. & 84.2 & 82.4 & 84.9 & 83.1 & 83.3 \\
\midrule
\multirow{4}{*}{ZS(CP)}
& C. & 99.9 & 99.9 & 99.9 & 99.9 & 99.9 \\
& L. & 70.9 & 69.9 & 70.4 & 68.9 & 69.0 \\
& S. & 74.9 & 76.7 & 77.2 & 77.9 & 77.9 \\
& V. & 85.9 & 85.2 & 86.0 & 84.9 & 85.0 \\
\midrule
\multirow{4}{*}{ZS(PC)}
& C. & 99.9 & 99.9 & 99.9 & 99.9 & 99.9 \\
& L. & 71.2 & 69.9 & 70.7 & 69.6 & 69.7 \\
& S. & 74.6 & 76.4 & 76.9 & 77.7 & 77.5 \\
& V. & 86.3 & 85.4 & 86.2 & 85.0 & 85.0 \\
\bottomrule
\end{tabular*}

\vspace{1em}

% ================================================================
% (b) Hidden dimension
% ================================================================
\begin{tabular*}{\linewidth}{@{\extracolsep{\fill}}cc*{7}{c}@{}}
\toprule
\multicolumn{9}{c}{\textbf{(b) Hidden dimension \(d_{\mathrm{mid}}\)}} \\
\midrule
Prompt & Domain
& 128 & 192 & 256 & 320 & 384 & 448 & 512 \\
\midrule
\multirow{4}{*}{ZS(C)}
& C. & 99.9 & 99.9 & 99.9 & 99.9 & 99.9 & 99.9 & 99.9 \\
& L. & 66.0 & 64.9 & 63.8 & 66.0 & 65.8 & 65.8 & 67.7 \\
& S. & 77.6 & 77.9 & 77.6 & 77.8 & 76.9 & 77.4 & 78.0 \\
& V. & 82.6 & 83.0 & 82.7 & 82.9 & 82.8 & 82.1 & 84.9 \\
\midrule
\multirow{4}{*}{ZS(CP)}
& C. & 99.9 & 99.9 & 99.9 & 99.9 & 99.9 & 99.9 & 99.9 \\
& L. & 70.0 & 68.9 & 67.6 & 69.2 & 69.4 & 69.7 & 70.4 \\
& S. & 77.4 & 78.0 & 78.7 & 78.1 & 77.5 & 77.6 & 77.2 \\
& V. & 85.4 & 85.6 & 84.8 & 85.3 & 85.3 & 84.9 & 86.0 \\
\midrule
\multirow{4}{*}{ZS(PC)}
& C. & 99.9 & 99.9 & 99.9 & 99.9 & 99.9 & 99.9 & 99.9 \\
& L. & 70.1 & 69.0 & 67.8 & 69.7 & 69.6 & 69.9 & 70.7 \\
& S. & 77.1 & 77.8 & 78.6 & 77.7 & 77.0 & 77.1 & 76.9 \\
& V. & 85.7 & 86.0 & 85.2 & 85.5 & 85.5 & 85.6 & 86.2 \\
\bottomrule
\end{tabular*}

\vspace{1em}

% ================================================================
% (c) Inference weight
% ================================================================
\begin{tabular*}{\linewidth}{@{\extracolsep{\fill}}cc*{7}{c}@{}}
\toprule
\multicolumn{9}{c}{\textbf{(c) Inference weight \(\alpha\)}} \\
\midrule
Prompt & Domain
& \(10^{-1.5}\)
& \(10^{-1}\)
& \(10^{-0.5}\)
& \(10^{0}\)
& \(10^{0.5}\)
& \(10^{1}\)
& \(10^{1.5}\) \\
\midrule
\multirow{4}{*}{ZS(C)}
& C. & 99.9 & 99.9 & 99.9 & 99.8 & 99.8 & 99.8 & 99.8 \\
& L. & 66.5 & 69.5 & 70.6 & 71.5 & 72.0 & 72.1 & 72.1 \\
& S. & 77.9 & 77.5 & 75.2 & 73.5 & 73.1 & 72.8 & 72.8 \\
& V. & 83.1 & 85.7 & 85.5 & 83.5 & 85.5 & 85.4 & 85.4 \\
\midrule
\multirow{4}{*}{ZS(CP)}
& C. & 99.9 & 99.9 & 99.9 & 99.9 & 99.8 & 99.8 & 99.8 \\
& L. & 70.4 & 70.4 & 70.7 & 71.7 & 72.2 & 72.3 & 72.2 \\
& S. & 77.1 & 77.1 & 75.7 & 74.4 & 73.7 & 73.4 & 73.3 \\
& V. & 86.0 & 86.2 & 85.9 & 85.8 & 85.7 & 85.7 & 85.7 \\
\midrule
\multirow{4}{*}{ZS(PC)}
& C. & 99.9 & 99.9 & 99.9 & 99.8 & 99.8 & 99.8 & 99.8 \\
& L. & 70.3 & 70.9 & 71.0 & 72.3 & 72.5 & 72.5 & 72.6 \\
& S. & 76.6 & 76.5 & 75.1 & 73.5 & 72.9 & 72.9 & 72.7 \\
& V. & 86.1 & 86.1 & 85.7 & 85.5 & 85.6 & 85.5 & 85.5 \\
\bottomrule
\end{tabular*}
\end{table}

\subsection{LLM Prompts Generation}
\label{subsec:llm-gen}

We employ ChatGPT-3.5 \cite{brown2020language} to generate the LLM-based prompts used in our comparative analysis of prompt sources. The procedure, illustrated in the following toolbox, involves prompting ChatGPT-3.5 to produce semantically rich, stylistically varied textual descriptions for specific class names. For each class, we generate 120 distinct samples. \cref{tab:llm-prompt-examples} shows representative examples of the generated prompts.

\begin{tcolorbox}[colback=gray!5, colframe=black!70, title=LLM Prompting Procedure (ChatGPT-3.5)]
\textbf{User:} I have some class names, such as \texttt{"dog"}, \texttt{"car"}, \texttt{"person"}, \texttt{"chair"}. I want you to generate various styles of these objects in different art-style pictures. Provide me with a list of different sentences that describe a scene containing the object, with different object attributes (such as object size, condition or age (if applicable), colors), different backgrounds the object may appear in; different image types (such as photo, painting, etc).

\vspace{1em}
\textbf{ChatGPT:} Absolutely, I'll provide you with a list of descriptive sentences for each of the class names \texttt{"dog"}, \texttt{"car"}, \texttt{"person"}, \texttt{"chair"}, incorporating various object attributes, backgrounds, and image types.

Each sentence will describe a unique scene:
\end{tcolorbox}

\begin{table}[tb]
\caption{
Representative examples of LLM-generated training prompts for selected classes. Each description was generated by ChatGPT-3.5 to provide varied attributes, contexts, and visual realizations.
}
\label{tab:llm-prompt-examples}
\centering
\begin{tabularx}{\textwidth}{@{}p{2cm}X@{}}
\toprule
\textbf{Class} & \textbf{Generated Prompt} \\
\midrule
\textbf{Bird} 
& A pair of vibrant macaws converse in a lush, tropical rainforest, depicted in a lively, exotic wildlife painting. \\
& A solitary eagle watches over a vast, rugged canyon at sunrise, portrayed in a majestic, wilderness landscape photograph. \\
\midrule
\textbf{Dog} 
& A sleek Whippet races in a competitive dog track, illustrated in a fast-paced, dynamic sports style. \\
& A sturdy and reliable English Bulldog watching over a small shop, its solid presence reassuring to the owner. \\
\midrule
\textbf{Car} 
& A quirky art car parades through the streets in a colorful festival, captured in a fun, expressive style illustration. \\
& A high-tech, autonomous car maneuvers through a smart city environment, portrayed in a futuristic, sci-fi digital art piece. \\
\midrule
\textbf{Chair} 
& A folding chair at an outdoor wedding, elegantly decorated and part of a beautiful ceremony. \\
& A high-end executive chair in a law firm, projecting authority and professionalism. \\
\midrule
\textbf{Person} 
& An energetic coach motivates a team on a sports field, illustrated in an inspiring, leadership-focused painting. \\
& A graceful figure skater glides across an ice rink, captured in a delicate, winter-themed pastel drawing. \\
\bottomrule
\end{tabularx}
\end{table}

\section{Experiments on Other CLIP Model Scales}
\label{sec:repeated_exps}

\subsection{Experiments on ViT-L/14}
\label{subsec:vitl14}

We refined the output dimension to align with the input dimension of 768. The chosen latent dimensions were 448 and 640 for PACS and VLCS, respectively, and 768 for both OfficeHome and DomainNet. The inference weighting $\alpha$ was set to 0.1 for PACS, 0.03 for VLCS, 0.14 for OfficeHome, and 0.2 for DomainNet. All other training configurations remained consistent with the ViT-B/16 experiments across each dataset. The training configuration for Im.Aug was set the same as CLAP for each dataset, with the inference weighting $\alpha$ being 0.1 for PACS and 0.03 for the other three datasets.

\Cref{tab:zs-vitl} showcases the zero-shot results for the ViT-L/14 model using four distinct prompts, following the protocol established for the ViT-B/16 experiments. These results demonstrate that CLAP is more efficient than Im.Aug in enhancing zero-shot performance. Moreover, \cref{tab:variance-vitl} illustrates that CLAP significantly reduces variations in zero-shot performance across different prompts, thereby confirming CLAP's performance improvements over CLIP across a range of model sizes. Detailed domain-level results are presented in \cref{tab:zs-vitl-dm}, offering an in-depth analysis. 

\begin{table}[tbp]
\caption{Zero-shot performance across four prompts ("C", "PC", "CP") and 1 noised prompts ("NC") with CLIP pre-trained ViT-L/14 model. CLAP demonstrates consistent gains in zero-shot performance across all datasets, validating its effectiveness.}
\label{tab:zs-vitl}
\centering
% \resizebox{\linewidth}{!}{
    \begin{tabular}{@{}ccccccc@{}}
    \toprule
    \multirow{2}{*}{Prompt} & \multirow{2}{*}{Method} & \multicolumn{5}{c}{Zero-shot performance,  avg. top-1 acc. (\%) (↑)}          \\ \cmidrule{3-7}
                            &                         & ~~PACS~~          & ~~VLCS~~          & OfficeHome    & DomainNet     & Overall       \\ \midrule
    \multirow{3}{*}{ZS(C)}  & CLIP                    & 97.6          & 77.1          & 85.9          & 63.2          & 80.9          \\
                            & Im.Aug                  & 98.3          & 78.5          & 86.0          & 63.4          & 81.6          \\
                            & CLAP                    & \textbf{98.5} & \textbf{80.7} & \textbf{87.5} & \textbf{64.2} & \textbf{82.7} \\ \midrule
    \multirow{3}{*}{ZS(CP)} & CLIP                    & 97.3          & 80.6          & 86.0          & 62.0          & 81.5          \\
                            & Im.Aug                  & 98.3          & 81.1          & 86.1          & 62.4          & 82.0          \\
                            & CLAP                    & \textbf{98.5} & \textbf{81.4} & \textbf{87.9} & \textbf{63.7} & \textbf{82.9} \\ \midrule
    \multirow{3}{*}{ZS(PC)} & CLIP                    & 98.4          & 81.7          & 86.5          & 63.5          & 82.5         \\
                            & Im.Aug                  & \textbf{98.6} & 81.9          & 86.6          & 63.7          & 82.7         \\
                            & CLAP                    & \textbf{98.6} & \textbf{82.2} & \textbf{88.0} & \textbf{64.5} & \textbf{83.3} \\ \midrule
    \multirow{3}{*}{ZS(NC)} & CLIP                    & 91.0          & 65.5          & 77.1          & 55.4          & 72.3          \\
                            & Im.Aug                  & 95.6          & 69.3          & 77.1          & 55.7          & 74.4          \\
                            & CLAP                    & \textbf{98.5} & \textbf{73.1} & \textbf{81.3} & \textbf{58.3} & \textbf{77.8} \\ \bottomrule
    \end{tabular}
    % }
\end{table}

\begin{table}[tbp]
\caption{CLAP reduces the variance in zero-shot performance across different prompts with a CLIP-pre-trained ViT-L/14 model.}
\label{tab:variance-vitl}
\centering
% \resizebox{\linewidth}{!}{
    \begin{tabular}{@{}ccccccc@{}}
    \toprule
    \multirow{2}{*}{Metric}  & \multirow{2}{*}{Method} & \multicolumn{5}{c}{Zero-shot variance, avg. top-1 acc. (\%)  (↓)} \\ \cmidrule{3-7}
                             &                         & ~~PACS~~          & ~~VLCS~~          & OfficeHome    & DomainNet     & Overall      \\ \midrule
    \multirow{3}{*}{$R$}     & CLIP                    & 1.0           & 4.6           & 0.6           & 1.5           & 1.9          \\
                             & Im.Aug                  & 0.3           & 3.4           & 0.6           & 1.3           & 1.4          \\
                             & CLAP                    & \textbf{0.1}  & \textbf{1.5}  & \textbf{0.4}  & \textbf{0.7}  & \textbf{0.7} \\ \midrule
    \multirow{3}{*}{$\delta$}& CLIP                    & 0.4           & 2.0           & 0.3           & 0.6           & 0.8          \\
                             & Im.Aug                  & 0.1           & 1.5           & 0.3           & 0.5           & 0.6          \\
                             & CLAP                    & \textbf{0.0}  & \textbf{0.6}  & \textbf{0.2}  & \textbf{0.3}  & \textbf{0.3} \\ \midrule
    \multirow{3}{*}{$\Delta_{(NC)}$} & CLIP            & 6.6           & 11.5          & 8.8           & 7.8           & 8.7          \\
                             & Im.Aug                  & 2.7           & 9.2           & 8.9           & 7.7           & 7.1          \\
                             & CLAP                    & \textbf{0.1}  & \textbf{7.7}  & \textbf{6.3}  & \textbf{5.9}  & \textbf{5.0} \\ \bottomrule
    \end{tabular}
% }
\end{table}

\begin{table}[tbp]
\caption{Domain-level zero-shot top-1 accuracy (\%) with the ViT-L/14 CLIP backbone. Panels (a) and (b) report results for the four inference prompt constructions.}
\label{tab:zs-vitl-dm}
\centering
\small
\setlength{\tabcolsep}{4pt}
\renewcommand{\arraystretch}{1.05}

% ================================================================
% (a) ZS(C) and ZS(CP)
% ================================================================
\begin{tabular*}{\linewidth}{@{\extracolsep{\fill}}ll*{6}{c}@{}}
\toprule
\multicolumn{8}{c}{\textbf{(a) ZS(C) and ZS(CP)}} \\
\midrule
\multirow{2}{*}{Dataset}
& \multirow{2}{*}{Domain}
& \multicolumn{3}{c}{ZS(C)}
& \multicolumn{3}{c}{ZS(CP)} \\
\cmidrule(lr){3-5}
\cmidrule(l){6-8}
&
& CLIP & Im.Aug & CLAP
& CLIP & Im.Aug & CLAP \\
\midrule

\multirow{4}{*}{PACS}
& A. & 97.2 & 98.0 & 98.8 & 96.8 & 98.0 & 98.5 \\
& C. & 99.5 & 99.6 & 99.8 & 98.3 & 99.6 & 99.7 \\
& P. & 99.9 & 100.0 & 100.0 & 99.4 & 99.5 & 100.0 \\
& S. & 93.8 & 95.7 & 95.5 & 94.8 & 96.0 & 95.7 \\
\midrule

\multirow{4}{*}{VLCS}
& C. & 99.9 & 99.9 & 99.9 & 99.9 & 99.9 & 99.9 \\
& L. & 57.4 & 60.1 & 64.3 & 71.3 & 71.6 & 72.6 \\
& S. & 71.0 & 72.4 & 74.4 & 66.2 & 67.4 & 66.8 \\
& V. & 80.0 & 81.6 & 84.3 & 85.2 & 85.7 & 86.2 \\
\midrule

\multirow{4}{*}{OfficeHome}
& A. & 86.2 & 86.3 & 87.7 & 85.7 & 86.2 & 88.1 \\
& C. & 73.3 & 73.4 & 75.7 & 73.8 & 73.4 & 76.0 \\
& P. & 92.0 & 91.8 & 93.6 & 92.3 & 92.4 & 94.3 \\
& R. & 92.2 & 92.7 & 93.0 & 92.2 & 92.4 & 93.4 \\
\midrule

\multirow{6}{*}{DomainNet}
& C. & 78.4 & 78.5 & 79.1 & 77.5 & 77.7 & 78.8 \\
& I. & 52.9 & 53.0 & 54.6 & 50.4 & 50.7 & 53.6 \\
& P. & 70.4 & 70.8 & 72.4 & 68.9 & 69.9 & 72.1 \\
& Q. & 21.5 & 21.6 & 22.5 & 20.6 & 20.9 & 21.7 \\
& R. & 85.8 & 85.9 & 85.9 & 85.3 & 85.5 & 85.7 \\
& S. & 70.2 & 70.4 & 70.7 & 69.4 & 69.8 & 70.6 \\
\bottomrule
\end{tabular*}

\vspace{1em}

% ================================================================
% (b) ZS(PC) and ZS(NC)
% ================================================================
\begin{tabular*}{\linewidth}{@{\extracolsep{\fill}}ll*{6}{c}@{}}
\toprule
\multicolumn{8}{c}{\textbf{(b) ZS(PC) and ZS(NC)}} \\
\midrule
\multirow{2}{*}{Dataset}
& \multirow{2}{*}{Domain}
& \multicolumn{3}{c}{ZS(PC)}
& \multicolumn{3}{c}{ZS(NC)} \\
\cmidrule(lr){3-5}
\cmidrule(l){6-8}
&
& CLIP & Im.Aug & CLAP
& CLIP & Im.Aug & CLAP \\
\midrule

\multirow{4}{*}{PACS}
& A. & 98.7 & 98.8 & 98.9 & 85.6 & 91.6 & 98.6 \\
& C. & 99.5 & 99.6 & 99.7 & 95.9 & 98.1 & 99.6 \\
& P. & 99.9 & 100.0 & 99.9 & 91.1 & 97.5 & 99.9 \\
& S. & 95.4 & 95.9 & 95.8 & 91.5 & 95.2 & 95.8 \\
\midrule

\multirow{4}{*}{VLCS}
& C. & 99.9 & 99.9 & 99.9 & 87.5 & 87.9 & 94.4 \\
& L. & 71.7 & 72.0 & 72.6 & 53.8 & 59.7 & 60.7 \\
& S. & 69.9 & 70.4 & 69.9 & 55.9 & 60.5 & 62.9 \\
& V. & 85.1 & 85.3 & 86.4 & 65.0 & 69.3 & 74.3 \\
\midrule

\multirow{4}{*}{OfficeHome}
& A. & 87.0 & 87.0 & 87.8 & 78.1 & 77.1 & 80.7 \\
& C. & 73.1 & 73.5 & 76.0 & 65.9 & 66.3 & 70.6 \\
& P. & 92.9 & 92.8 & 94.1 & 80.7 & 81.0 & 86.8 \\
& R. & 93.1 & 93.3 & 93.9 & 83.8 & 84.0 & 86.9 \\
\midrule

\multirow{6}{*}{DomainNet}
& C. & 79.4 & 79.4 & 79.7 & 70.0 & 70.4 & 72.8 \\
& I. & 51.7 & 52.0 & 53.9 & 45.3 & 45.2 & 48.8 \\
& P. & 69.9 & 70.6 & 72.7 & 59.9 & 60.3 & 64.8 \\
& Q. & 22.6 & 22.8 & 22.9 & 17.9 & 18.4 & 20.2 \\
& R. & 86.3 & 86.4 & 86.2 & 77.5 & 77.5 & 78.7 \\
& S. & 71.0 & 71.3 & 71.5 & 62.0 & 62.2 & 64.6 \\
\bottomrule
\end{tabular*}

\end{table}

\subsection{Experiments on ResNet50x16}
\label{subsec:rn50x16}

To validate our approach on different model structures, we repeated zero-shot experiments on the ResNet50x16 model pre-trained with CLIP. Since the output dimension of CLIP is the same as ViT-B/16, we used the same training configuration as ViT-B/16 for training Im.Aug and CLAP. For inference, we refined the weighting coefficient $\alpha$ to 0.1, 1, 0.03, and 0.1 for Im.Aug, and 0.03, 0.2, 0.06, and 0.1 for CLAP, for PACS, VLCS, OfficeHome, and DomainNet respectively.

\Cref{tab:zs-rn50} showcases the zero-shot results for ResNet50x16 model across different prompts, substantiating that CLAP is more effective than Im.Aug in refining CLIP features. Moreover, \cref{tab:variance-rn50} illustrates that both Im.Aug and CLAP reduce variations in zero-shot performance across different prompts, with the improvement of CLAP being more significant. The results validate our approach across different model scales, including both ViT-based and CNN-based structures. Domain-level results are detailed in \cref{tab:zs-rn50x16-dm}.

\begin{table}[tbp]
\caption{Zero-shot performance with CLIP pre-trained ResNet50x16 model. CLAP demonstrates consistent enhancement across all datasets, validating its effectiveness.}
\label{tab:zs-rn50}
\centering
% \resizebox{\linewidth}{!}{
    \begin{tabular}{@{}ccccccc@{}}
    \toprule
    \multirow{2}{*}{Prompt} & \multirow{2}{*}{Method} & \multicolumn{5}{c}{Zero-shot performance,  avg. top-1 acc. (\%) (↑)}          \\ \cmidrule{3-7}
                            &                         & ~~PACS~~          & ~~VLCS~~          & OfficeHome    & DomainNet     & Overall       \\ \midrule
    \multirow{3}{*}{ZS(C)}  & CLIP                    & 96.1         & 70.4          & 80.4          & 57.1          & 76.0          \\
                            & Im.Aug                  & 96.4         & 74.7          & 80.4          & 57.1          & 77.2          \\
                            & CLAP                    & \textbf{97.0} & \textbf{79.9} & \textbf{81.6} & \textbf{58.0} & \textbf{79.1} \\ \midrule
    \multirow{3}{*}{ZS(CP)} & CLIP                    &  95.0         & 73.5          & 79.0          & 56.1          & 75.9          \\
                            & Im.Aug                  & 95.7          & 75.8          & 79.3          & 56.5          & 76.8          \\
                            & CLAP                    & \textbf{96.7} & \textbf{80.3} & \textbf{79.9} & \textbf{57.4} & \textbf{78.6} \\ \midrule
    \multirow{3}{*}{ZS(PC)} & CLIP                    & 96.5          & 78.4          & 81.7         & 57.1          & 78.4          \\
                            & Im.Aug                  & 97.0          & 79.8          & 81.8          & 57.4          & 79.0          \\
                            & CLAP                    & \textbf{96.8} & \textbf{80.1} & \textbf{82.5} & \textbf{58.2} & \textbf{79.4} \\ \midrule
    \multirow{3}{*}{ZS(NC)} & CLIP                    & 86.4          & 61.2          & 69.3          & 48.2          & 66.3         \\
                            & Im.Aug                  & 88.3    & 71.3          & 69.5          & 48.7          & 69.4         \\
                            & CLAP                    & \textbf{94.9} & \textbf{80.1} & \textbf{71.9} & \textbf{50.6} & \textbf{74.4} \\ \bottomrule

    \end{tabular}
    % }
\end{table}

\begin{table}[tbp]
\caption{CLAP consistently reduces variances in zero-shot performance across different prompts with CLIP pre-trained ResNet50x16 model, validating its effectiveness.}
\label{tab:variance-rn50}
\centering
% \resizebox{\linewidth}{!}{
    \begin{tabular}{@{}ccccccc@{}}
    \toprule
    \multirow{2}{*}{Metric}  & \multirow{2}{*}{Method} & \multicolumn{5}{c}{Zero-shot variance, avg. top-1 acc. (\%)  (↓)} \\ \cmidrule{3-7}
                             &                         & ~~PACS~~          & ~~VLCS~~          & OfficeHome    & DomainNet     & Overall      \\ \midrule
    \multirow{3}{*}{$R$}     & CLIP                    & 1.5           & 8.0          & 2.7           & 1.1           & 3.3          \\
                             & Im.Aug                  &  1.3          &  5.1          & \textbf{2.5}           & 0.9           & 2.4          \\
                             & CLAP                    & \textbf{0.3}  & \textbf{0.4}  & 2.6  & \textbf{0.8}  & \textbf{1.0} \\ \midrule
    \multirow{3}{*}{$\delta$}& CLIP                    & 0.6           & 3.3           & 1.1           & 0.5           & 1.4          \\
                             & Im.Aug                  & 0.5           & 2.2           & \textbf{1.0}           & 0.4           &  1.0         \\
                             & CLAP                    & \textbf{0.1}  & \textbf{0.2}  & 1.1  & \textbf{0.3}  & \textbf{0.4} \\ \midrule
    \multirow{3}{*}{$\Delta_{(NC)}$} & CLIP            & 9.7           & 9.3          & 11.1           & 8.9           & 9.7          \\
                             & Im.Aug                  & 8.1           & 3.5           & 10.9          & 8.5           & 7.7         \\
                             & CLAP                    & \textbf{2.1}  & \textbf{-0.1}  & \textbf{9.7}  & \textbf{7.5}  & \textbf{4.8} \\ \bottomrule
    \end{tabular}
% }
\end{table}

\begin{table}[tbp]
\caption{Domain-level zero-shot top-1 accuracy (\%) with the ResNet50x16 CLIP backbone. Panels (a) and (b) report results for the four inference prompt constructions.}
\label{tab:zs-rn50x16-dm}
\centering
\small
\setlength{\tabcolsep}{4pt}
\renewcommand{\arraystretch}{1.05}

% ================================================================
% (a) ZS(C) and ZS(CP)
% ================================================================
\begin{tabular*}{\linewidth}{@{\extracolsep{\fill}}ll*{6}{c}@{}}
\toprule
\multicolumn{8}{c}{\textbf{(a) ZS(C) and ZS(CP)}} \\
\midrule
\multirow{2}{*}{Dataset}
& \multirow{2}{*}{Domain}
& \multicolumn{3}{c}{ZS(C)}
& \multicolumn{3}{c}{ZS(CP)} \\
\cmidrule(lr){3-5}
\cmidrule(l){6-8}
&
& CLIP & Im.Aug & CLAP
& CLIP & Im.Aug & CLAP \\
\midrule

\multirow{4}{*}{PACS}
& A. & 95.7 & 95.8 & 97.2 & 93.7 & 95.5 & 96.5 \\
& C. & 98.3 & 98.2 & 99.0 & 98.1 & 98.8 & 99.0 \\
& P. & 98.9 & 98.6 & 99.9 & 98.4 & 97.8 & 99.8 \\
& S. & 91.5 & 93.1 & 91.9 & 89.7 & 90.8 & 91.5 \\
\midrule

\multirow{4}{*}{VLCS}
& C. & 96.8 & 97.1 & 99.3 & 99.7 & 99.4 & 99.3 \\
& L. & 53.4 & 60.8 & 65.9 & 51.6 & 58.9 & 67.3 \\
& S. & 63.2 & 70.9 & 69.5 & 68.0 & 72.9 & 69.5 \\
& V. & 68.4 & 70.1 & 85.2 & 74.5 & 72.1 & 85.3 \\
\midrule

\multirow{4}{*}{OfficeHome}
& A. & 82.2 & 82.5 & 83.5 & 79.7 & 79.9 & 80.6 \\
& C. & 63.0 & 62.9 & 64.7 & 61.7 & 62.2 & 62.8 \\
& P. & 88.2 & 87.9 & 89.0 & 87.4 & 87.5 & 88.5 \\
& R. & 88.1 & 88.2 & 89.1 & 87.3 & 87.5 & 87.6 \\
\midrule

\multirow{6}{*}{DomainNet}
& C. & 69.0 & 68.9 & 69.6 & 68.6 & 68.6 & 69.4 \\
& I. & 51.0 & 51.1 & 52.7 & 48.2 & 49.0 & 50.6 \\
& P. & 65.2 & 65.6 & 66.5 & 63.7 & 64.4 & 65.6 \\
& Q. & 11.8 & 11.9 & 12.7 & 12.3 & 12.6 & 13.1 \\
& R. & 82.1 & 82.2 & 83.1 & 81.6 & 81.8 & 82.6 \\
& S. & 63.2 & 63.1 & 63.6 & 62.0 & 62.4 & 63.4 \\
\bottomrule
\end{tabular*}

\vspace{1em}

% ================================================================
% (b) ZS(PC) and ZS(NC)
% ================================================================
\begin{tabular*}{\linewidth}{@{\extracolsep{\fill}}ll*{6}{c}@{}}
\toprule
\multicolumn{8}{c}{\textbf{(b) ZS(PC) and ZS(NC)}} \\
\midrule
\multirow{2}{*}{Dataset}
& \multirow{2}{*}{Domain}
& \multicolumn{3}{c}{ZS(PC)}
& \multicolumn{3}{c}{ZS(NC)} \\
\cmidrule(lr){3-5}
\cmidrule(l){6-8}
&
& CLIP & Im.Aug & CLAP
& CLIP & Im.Aug & CLAP \\
\midrule

\multirow{4}{*}{PACS}
& A. & 95.7 & 96.7 & 96.8 & 81.2 & 84.0 & 94.5 \\
& C. & 98.6 & 98.7 & 98.9 & 92.3 & 93.2 & 98.0 \\
& P. & 99.8 & 99.9 & 99.9 & 85.3 & 87.3 & 95.2 \\
& S. & 91.8 & 92.9 & 91.6 & 86.9 & 88.6 & 92.1 \\
\midrule

\multirow{4}{*}{VLCS}
& C. & 99.7 & 99.6 & 99.4 & 75.6 & 89.3 & 99.4 \\
& L. & 59.5 & 68.1 & 66.8 & 54.1 & 60.6 & 67.0 \\
& S. & 72.9 & 73.7 & 69.0 & 52.0 & 66.7 & 69.5 \\
& V. & 81.7 & 78.0 & 85.2 & 63.1 & 68.5 & 84.5 \\
\midrule

\multirow{4}{*}{OfficeHome}
& A. & 82.0 & 82.4 & 83.4 & 67.7 & 68.9 & 72.1 \\
& C. & 65.4 & 65.3 & 66.1 & 54.6 & 55.0 & 56.8 \\
& P. & 90.0 & 89.9 & 90.6 & 75.4 & 75.3 & 78.2 \\
& R. & 89.2 & 89.5 & 89.7 & 79.5 & 78.9 & 80.3 \\
\midrule

\multirow{6}{*}{DomainNet}
& C. & 69.9 & 70.0 & 70.4 & 59.5 & 60.1 & 61.4 \\
& I. & 48.2 & 48.9 & 50.7 & 41.2 & 41.6 & 44.3 \\
& P. & 65.4 & 65.9 & 67.0 & 53.5 & 54.3 & 56.8 \\
& Q. & 11.8 & 12.2 & 12.7 & 9.3 & 9.7 & 11.0 \\
& R. & 83.3 & 83.4 & 83.8 & 72.9 & 73.0 & 74.7 \\
& S. & 63.9 & 63.8 & 64.6 & 53.1 & 53.4 & 55.3 \\
\bottomrule
\end{tabular*}

\end{table}

\section{Discussion}
\label{sec:dicussison}

\subsection{Rationale behind CLAP's Foundation on CLIP}
\label{subsec:rationale}

The primary challenge in cross-modal transferability lies in the significant domain gap between text and image data, which typically hinders the direct application of models trained in one modality to another. For a causal explanation, despite the consistency of the content variable that dictates the object label across modalities, the generative processes from latent variables to observations inherent to each modality differ markedly. The CLIP model, trained on a comprehensive dataset of image-text pairs with a symmetric InfoNCE loss, significantly ameliorates this issue. By aligning the features of text and images into similar patterns, it facilitates leveraging a network trained atop the CLIP encoder of one modality as a viable proxy for the other. Consequently, this allows for the direct application of the disentangled network trained in the text modality atop CLIP's image encoder to refine representations.

\subsection{Impact of Image and Text Augmentations}
\label{subsec:impact_augmentations}

Identifying pure content factors poses a significant challenge. This difficulty primarily arises from the need for finding effective augmentations of observational data to alter style factors significantly while preserving content integrity. 

Through the cross-modal alignment provided by CLIP, we discovered that disentangling in one modality can seamlessly improve representations in both modalities. The impact of image augmentations has been well-explored and found effective at preserving content, but traditional methods do not impose sufficient changes to remove all style information. Our exploration of text augmentations reveals that the logical structure of text and the relative ease of implementing style changes can have a significant impact on achieving disentanglement. However, more efficient methods are worthy of exploration. 

A promising direction for future research is to explore efficient combinations of both modalities to enhance disentangled semantics. As each modality has its unique advantages—Text data recapitulates properties well since it is pre-processed by human intelligence, while image data is more precise in depicting the exact same objects or events due to its more detailed nature—the impact of combining augmentations of both modalities could be substantial. 

\end{document}